%% file: main.tex
\documentclass[onecolumn]{article}

\usepackage[utf8]{inputenc}
\usepackage[T1]{fontenc}
\usepackage{amsmath,amssymb,amsfonts}
\usepackage{graphicx}
\usepackage{booktabs}
\usepackage{natbib}
\usepackage[margin=1in]{geometry}
\usepackage{caption}
\usepackage{subcaption}
\usepackage{hyperref}
\usepackage{siunitx}
\usepackage{float}
\usepackage{xcolor}
\usepackage{multirow}

\title{\textbf{Causal Circuit Tracing Reveals Distinct Computational\\Architectures in Single-Cell Foundation Models:\\Inhibitory Dominance, Biological Coherence, and\\Cross-Model Convergence}}

\author{Ihor Kendiukhov\\
Department of Computer Science\\
University of T\"ubingen\\
T\"ubingen, Germany\\
\texttt{kendiukhov@gmail.com}}

\date{}

\begin{document}

\maketitle

\begin{abstract}
\textbf{Motivation:} Sparse autoencoders (SAEs) decompose foundation model activations into interpretable features, but causal feature-to-feature interactions across network depth remain unknown for biological foundation models.

\noindent\textbf{Results:} We introduce causal circuit tracing---ablating SAE features and measuring downstream responses---and apply it to Geneformer V2-316M and scGPT whole-human across four conditions (96,892 edges, 80,191 forward passes). Both models show ${\sim}53\%$ biological coherence and 65--89\% inhibitory dominance, invariant to architecture and cell type. scGPT produces stronger effects (mean $|d|{=}1.40$ vs.\ 1.05) with more balanced dynamics. Cross-model consensus yields 1,142 conserved domain pairs (10.6$\times$ enrichment, $p < 0.001$). Disease-associated domains are 3.59$\times$ more likely to be consensus. Gene-level CRISPRi validation shows 56.4\% directional accuracy, confirming co-expression rather than causal encoding.

\noindent\textbf{Availability and implementation:} \url{https://github.com/Biodyn-AI/bio-sae-circuits} (Python). Companion SAE atlases: \url{https://github.com/Biodyn-AI/bio-sae}.

\noindent\textbf{Contact:} \texttt{kendiukhov@gmail.com}

\noindent\textbf{Supplementary information:} Supplementary data are available at \textit{Bioinformatics} online.
\end{abstract}


\section{Introduction}

Sparse autoencoders~\citep{sharkey2022taking,cunningham2023sparse,bricken2023monosemanticity} address the superposition hypothesis~\citep{elhage2022superposition} by decomposing model activations into overcomplete dictionaries of monosemantic features~\citep{olshausen1997sparse}. Scaling these methods to large models~\citep{gao2024scaling,templeton2024scaling} has revealed interpretable features in language models; we recently extended this approach to biology.

We presented comprehensive SAE feature atlases for Geneformer V2-316M~\citep{theodoris2023transfer} and scGPT whole-human~\citep{cui2024scgpt}---two single-cell foundation models~\citep{yang2022scbert,hao2024large}---in a companion study~\citep{kendiukhov2025sae_atlas}, systematically characterizing over 107,000 features across 30 layers. That work established that both models encode rich biological knowledge---pathway membership, protein interactions, functional modules, hierarchical abstraction---but minimal causal regulatory logic. The atlases revealed \emph{what} features exist and \emph{where} they are; they did not reveal \emph{how} features causally interact across network depth.

Understanding feature-to-feature causal relationships is critical for mechanistic interpretability~\citep{bereska2024mechanistic}. Causal intervention methods---including activation patching~\citep{vig2020causal,meng2022locating}, causal abstraction~\citep{geiger2021causal}, and circuit discovery~\citep{wang2023interpretability,marks2024sparse}---have proven powerful for understanding information flow in language models, but have not been applied at the SAE feature level in biological foundation models. While co-activation statistics (pointwise mutual information) reveal which features tend to fire together~\citep{manning1999foundations}, they cannot distinguish correlation from causation or determine the direction and magnitude of information flow. Single-feature causal patching~\citep{kendiukhov2025sae_atlas} demonstrated that individual features have specific downstream effects on output logits (median 2.36$\times$ specificity), but this probed feature$\to$output relationships rather than the internal feature$\to$feature circuits that constitute the model's computational graph.

Here we introduce \textbf{causal feature-to-feature circuit tracing}: extending the logic of activation patching~\citep{vig2020causal,geiger2021causal} and sparse feature circuits~\citep{marks2024sparse} to the biological domain by systematically ablating source SAE features~\citep{cunningham2023sparse,bricken2023monosemanticity} at one layer and measuring how all downstream SAE features across subsequent layers change. This reveals the directed computational graph---the wiring diagram of biological information processing---that each model uses to transform gene-level input into contextual predictions. We apply this method to both Geneformer and scGPT under four experimental conditions, enabling a controlled decomposition of how circuit properties depend on model architecture, SAE training data, and input cell type.

\section{Results}

\subsection{Causal circuit tracing reveals dense, predominantly inhibitory computational graphs}
\label{sec:circuits}

We performed causal circuit tracing on Geneformer V2-316M using K562-only SAEs (all 18 layers, 4,608 features each) with 200 K562 control cells from the Replogle CRISPRi dataset~\citep{replogle2022mapping}. For each of 120 source features (30 well-annotated features at each of layers 0, 5, 11, and 15; hereafter L0, L5, L11, L15), we ablated the feature at its source layer and measured the resulting change in all downstream SAE feature activations across all subsequent layers (Methods).

\begin{figure}[H]
\centering
\includegraphics[width=\textwidth]{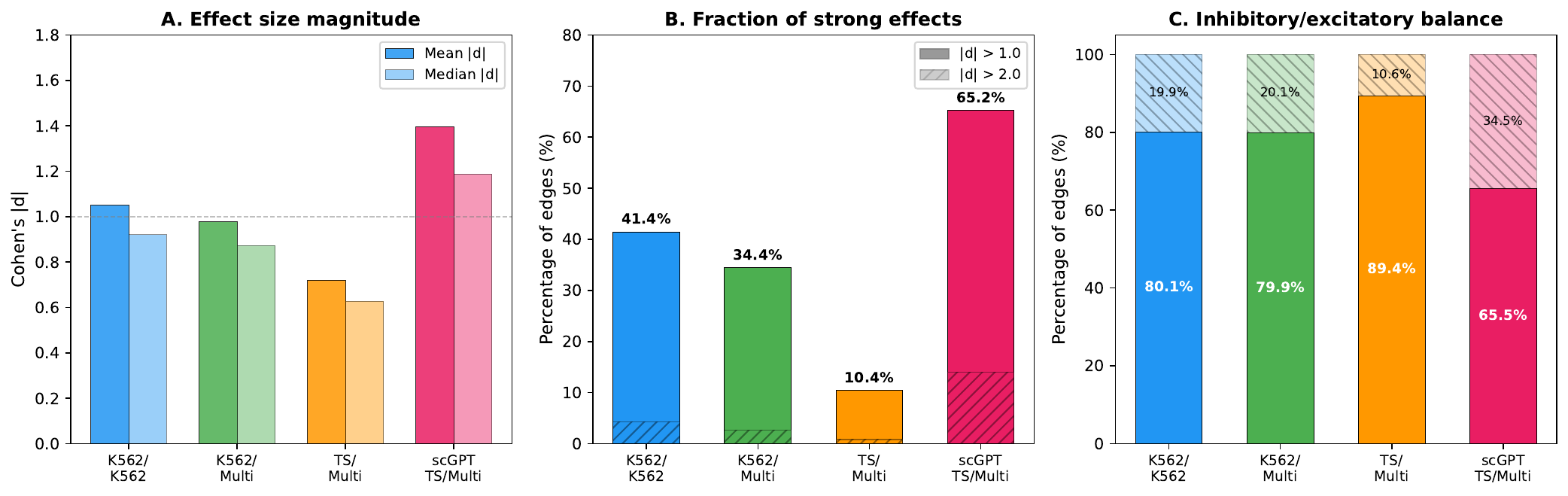}
\caption{\textbf{Causal effect size comparison across four experimental conditions.} \textbf{(A)}~Mean and median Cohen's $|d|$ for causal edges. scGPT produces the strongest individual effects ($|d|{=}1.40$), while Tabula Sapiens cells through Geneformer produce the weakest ($|d|{=}0.72$). The dashed line marks $|d|{=}1.0$ (strong effect threshold). \textbf{(B)}~Fraction of edges exceeding strong ($|d|{>}1.0$) and very strong ($|d|{>}2.0$) thresholds. scGPT leads with 65.2\% strong edges. \textbf{(C)}~Inhibitory/excitatory balance. All conditions are predominantly inhibitory (65--89\%), with Tabula Sapiens cells producing the most inhibitory circuits (89.4\%) and scGPT the most balanced (65.5\%).}
\label{fig:effect_sizes}
\end{figure}

The aggregate circuit graph contains \textbf{52,116 significant causal edges} ($|d| > 0.5$, consistency $> 0.7$), connecting 120 source features to 26,338 unique target features (Table~\ref{tab:aggregate}). Each source feature causally influences 615--2,459 downstream features on average (Table~\ref{tab:perlayer}), demonstrating that the model's computational graph is dense rather than sparse. Effect sizes are substantial: mean $|d| = 1.05$, median 0.92, with 41.4\% of edges exceeding $|d| > 1.0$ and 4.3\% exceeding $|d| > 2.0$ (Figure~\ref{fig:effect_sizes}A).

A striking property is the dominance of \textbf{inhibitory edges}: 80.1\% of causal edges have negative sign, meaning that ablating a source feature \emph{reduces} downstream feature activations (Figure~\ref{fig:effect_sizes}C). This implies that features predominantly encode necessary information---removing a feature causes downstream features that depend on it to lose activation---rather than redundant information (where removal would free capacity and increase other features). The 20\% excitatory fraction reflects disinhibition: ablating some features releases downstream features from suppression.

\begin{table}[t]
\centering
\caption{\textbf{Aggregate circuit statistics across four experimental conditions.} K562/K562: K562 cells with K562-only SAEs (all 18 layers). K562/Multi: K562 cells with multi-tissue SAEs (4 layers). TS/Multi (GF): Tabula Sapiens cells with multi-tissue SAEs via Geneformer. TS/Multi (scGPT): Tabula Sapiens cells with scGPT SAEs (all 12 layers).}
\label{tab:aggregate}
\smallskip
\begin{tabular}{lrrrr}
\toprule
Metric & K562/K562 & K562/Multi & TS/Multi (GF) & TS/Multi (scGPT) \\
\midrule
Model & Geneformer & Geneformer & Geneformer & scGPT \\
Layers with SAEs & 18 (all) & 4 (subset) & 4 (subset) & 12 (all) \\
Features/layer & 4,608 & 4,608 & 4,608 & 2,048 \\
Source features & 120 & 90 & 90 & 90 \\
Input cells & 200 K562 & 200 K562 & 200 TS & 200 TS \\
Forward passes & 24,776 & 18,465 & 18,455 & 18,495 \\
Compute time & 7.5 hr & 3.5 hr & 3.2 hr & 37 min \\
\midrule
Total edges & 52,116 & 8,298 & 5,098 & 31,380 \\
Target features & 26,338 & 4,171 & 2,962 & 1,960 \\
Target coverage & 31.9\% & --- & --- & 95.7\% \\
Mean $|d|$ & 1.05 & 0.98 & 0.72 & \textbf{1.40} \\
Median $|d|$ & 0.92 & 0.87 & 0.63 & \textbf{1.19} \\
$|d| > 1.0$ (\%) & 41.4 & 34.4 & 10.4 & \textbf{65.2} \\
Inhibitory (\%) & 80.1 & 79.9 & \textbf{89.4} & 65.5 \\
Shared ontology (\%) & 52.9 & \textbf{68.8} & 68.5 & 53.0 \\
\bottomrule
\end{tabular}
\end{table}

\begin{table}[t]
\centering
\caption{\textbf{Per-source-layer circuit statistics for Geneformer K562/K562.} Each source layer contributes 30 features. Downstream layers indicates the number of subsequent layers with SAEs available for measurement.}
\label{tab:perlayer}
\smallskip
\begin{tabular}{lrrrrrr}
\toprule
Source & Downstream & Passes & Time & Total Edges & Avg/Feature & Range \\
\midrule
L0 & 17 (L1--L17) & 6,176 & 150 min & 73,769 & 2,459 & 382--8,028 \\
L5 & 12 (L6--L17) & 6,200 & 140 min & 41,684 & 1,389 & 533--4,703 \\
L11 & 6 (L12--L17) & 6,200 & 86 min & 30,981 & 1,033 & 321--3,536 \\
L15 & 2 (L16--L17) & 6,200 & 71 min & 18,438 & 615 & 79--3,257 \\
\bottomrule
\end{tabular}
\end{table}

\subsection{Hub features and convergent integration layers}
\label{sec:hubs}

The circuit graph has highly asymmetric degree distributions. \textbf{Hub features at early layers} act as broadcast nodes, each causally influencing thousands of downstream features (Table~\ref{tab:hubs_gf}). The top hub---L0\_F2905 (Golgi Organization)---has an out-degree of 8,028, meaning its ablation significantly affects over 8,000 features across all 17 downstream layers (Supplementary Note~5). Other top hubs encode RNA processing (Supplementary Note~6), growth factor response, cholesterol biosynthesis (Supplementary Note~4), and RNA splicing---all broad cellular programs whose disruption would be expected to propagate widely.

Conversely, late-layer features show high \textbf{in-degree}, receiving causal input from many upstream sources. Layer~16 features achieve in-degrees of 88--93 (from a possible 120 source features), suggesting L16 serves as a convergent integration layer where diverse upstream biological computations are consolidated before the final output layer.

\begin{table}[t]
\centering
\caption{\textbf{Top hub features in Geneformer K562/K562 circuits.} Left: highest out-degree features (most downstream targets). Right: highest in-degree features (most upstream sources).}
\label{tab:hubs_gf}
\smallskip
\begin{tabular}{llr|lr}
\toprule
Feature & Biology & Out-Deg & Feature & In-Deg \\
\midrule
L0\_F2905 & Golgi Organization & 8,028 & L16\_F2818 & 93 \\
L0\_F2982 & RNA Methylation & 6,921 & L16\_F1691 & 89 \\
L0\_F1568 & Growth Factor Response & 6,006 & L16\_F4354 & 89 \\
L0\_F3402 & Cholesterol Biosynthesis & 5,096 & L16\_F1375 & 88 \\
L0\_F4201 & RNA Splicing & 4,782 & L16\_F1057 & 88 \\
\bottomrule
\end{tabular}
\end{table}

\subsection{Circuits encode interpretable biological cascades}
\label{sec:biology}

To assess whether causal circuits reflect meaningful biology, we tested whether source and target features of each edge share ontology annotations~\citep{ashburner2000go,kanehisa2000kegg,jassal2020reactome,szklarczyk2023string,han2018trrust}. Of 31,176 edges where both source and target have at least one annotation, \textbf{16,507 (52.9\%) share at least one ontology term} (GO Biological Process, KEGG, Reactome, STRING, or TRRUST). This means over half of the model's computational pathways connect biologically related features---far above what would be expected by chance given the thousands of possible ontology terms (Figure~\ref{fig:coherence}A).

\begin{figure}[H]
\centering
\includegraphics[width=\textwidth]{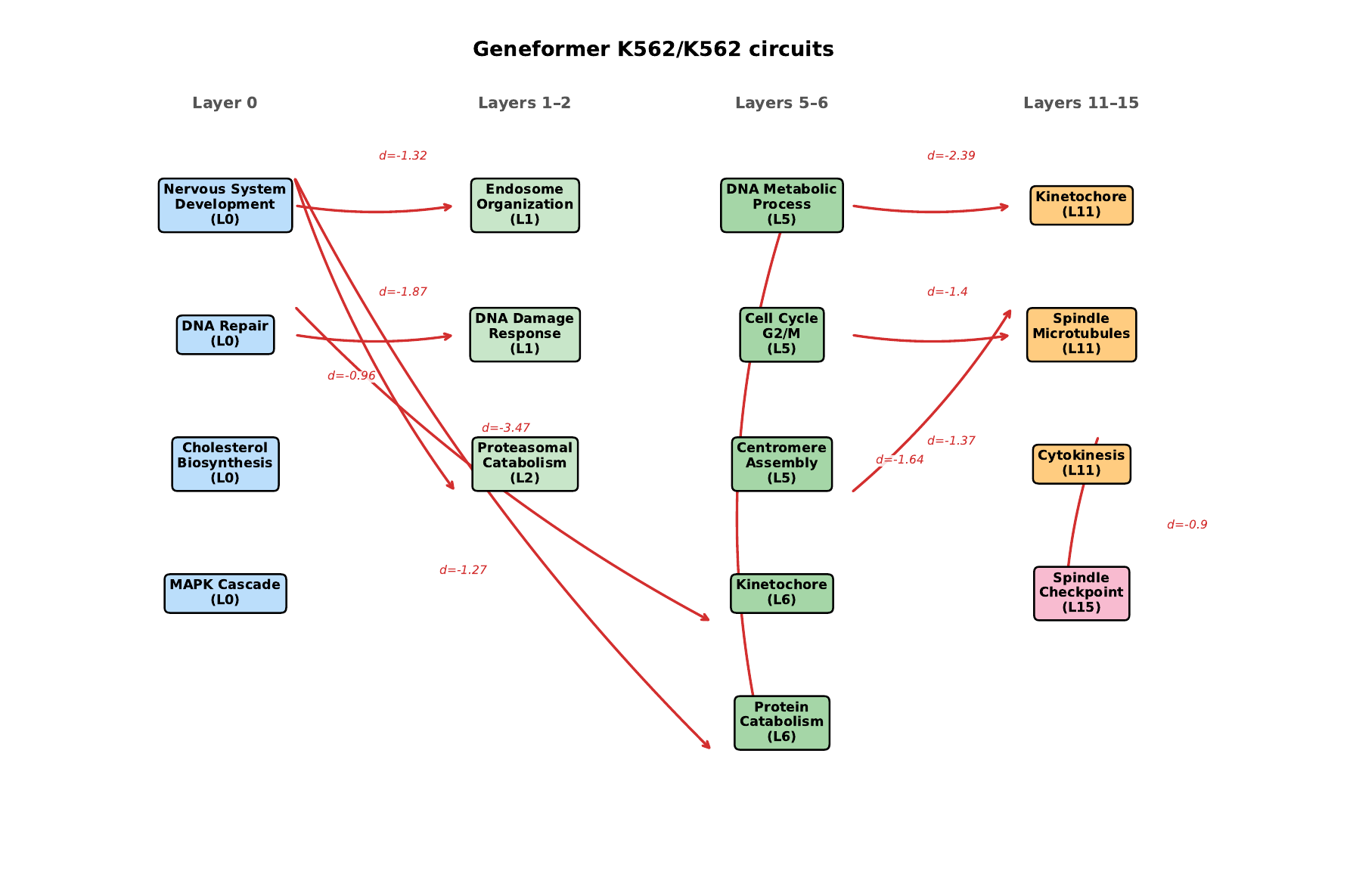}
\caption{\textbf{Geneformer K562/K562 biological circuit architecture.} 16 features across layers 0--15 connected by 10 significant causal edges. Two main circuit families emerge: (i)~the neurodevelopment$\to$proteostasis pathway (L0 Nervous System Development driving L1 Endosome Organization, L2 Proteasomal Catabolism, and L6 Protein Catabolism), and (ii)~the DNA damage$\to$mitotic apparatus cascade (L0 DNA Repair driving L6 Kinetochore at $d = -3.47$, then L5 DNA Metabolic driving L11 Kinetochore at $d = -2.39$).}
\label{fig:circuits}
\end{figure}

Specific circuits are directly interpretable as known biological cascades (Figure~\ref{fig:circuits}). We highlight several categories of biological circuits discovered across both models.

\subsubsection{DNA damage response cascades}

The most prominent circuit family in Geneformer involves the DNA damage response (DDR). An L0 DNA Repair feature (F3717) causally drives an L1 DNA Damage Response feature ($d = -1.87$, 113 shared ontology terms), which reflects the biological reality that DNA repair machinery depends on upstream damage detection signals~\citep{ciccia2010ddr,jackson2009ddr}. This same L0 repair feature produces an exceptionally strong connection to an L6 Kinetochore feature ($d = -3.47$), recapitulating the known link between DNA damage and mitotic checkpoint activation: cells with unresolved DNA damage activate kinetochore-associated checkpoint proteins to arrest cell division~\citep{musacchio2007spindle,malumbres2009cell}.

\begin{figure}[H]
\centering
\includegraphics[width=\textwidth]{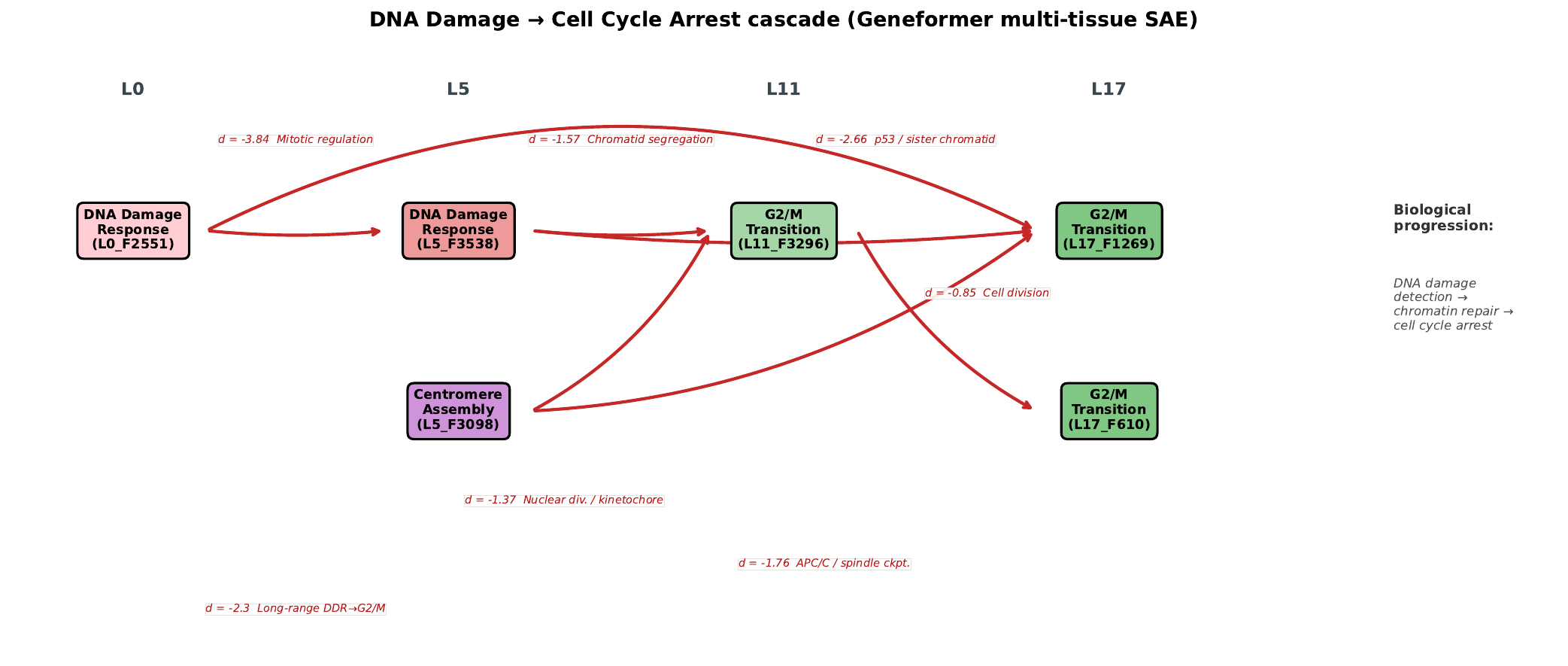}
\caption{\textbf{DNA damage response cascade in Geneformer multi-tissue circuits.} A biologically coherent multi-layer circuit progresses from DNA damage detection (L0) through checkpoint activation (L5) to cell cycle arrest (L11, L17). The L5 Centromere Assembly feature provides a parallel pathway through kinetochore organization. The L0 source directly connects to L17 targets ($d = -2.30$), creating a ``skip connection'' alongside the sequential cascade. Shared biological annotations (text on edges) confirm that each connection reflects established molecular pathways.}
\label{fig:cascade}
\end{figure}

In the multi-tissue SAE circuits, a three-layer DDR cascade emerges with even greater clarity (Figure~\ref{fig:cascade}). L0 DNA Damage Response (F2551) drives L5 DNA Damage Response (F3538, $d = -3.84$), sharing 72 ontology terms including mitotic regulation and spindle checkpoint. L5 DNA Damage Response then drives L17 G2/M Transition (F1269, $d = -2.66$), sharing 75 terms including sister chromatid separation and p53 signaling~\citep{ciccia2010ddr}. Remarkably, the L0 source also directly drives the L17 target ($d = -2.30$, 65 shared terms), revealing a ``skip connection'' in the biological circuit that bypasses the intermediate layer. This cascade represents the complete biological progression: \emph{DNA damage detection $\to$ checkpoint activation $\to$ cell cycle arrest}---precisely the sequence known from decades of cell biology research~\citep{jackson2009ddr,malumbres2009cell}.

\begin{figure}[H]
\centering
\includegraphics[width=\textwidth]{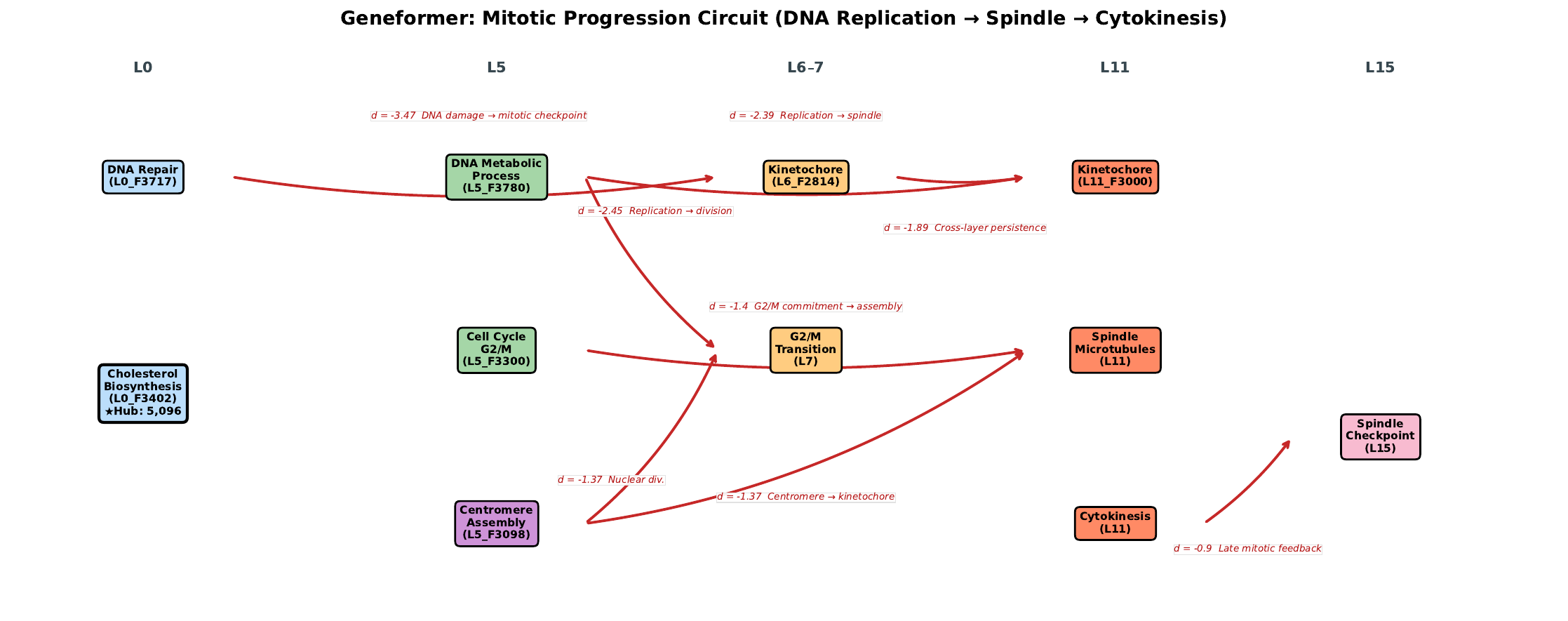}
\caption{\textbf{Geneformer: Mitotic progression circuit from DNA replication to cytokinesis.} Centromere assembly drives spindle checkpoint and G2/M transition across layers 0--15, terminating with cytokinesis feeding back to spindle checkpoint. See Supplementary Note~1 for full description.}
\label{fig:mitotic}
\end{figure}

\begin{figure}[H]
\centering
\includegraphics[width=\textwidth]{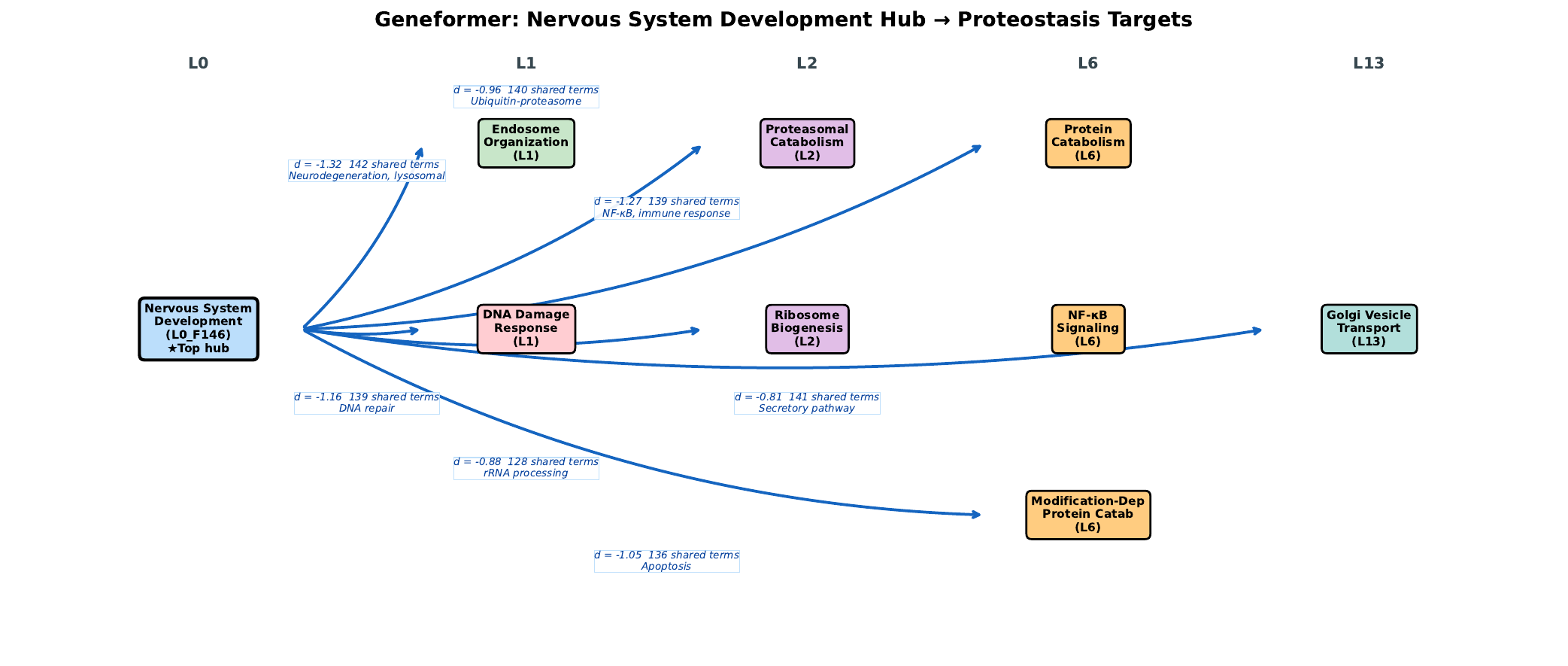}
\caption{\textbf{Geneformer: Nervous System Development as a computational hub.} L0 feature F146 drives 7 targets across layers 1--13 spanning proteostasis, cellular transport, and immune signaling (128--142 shared terms per edge). See Supplementary Note~2.}
\label{fig:neuro_hub}
\end{figure}

Additional circuit families include mitotic apparatus assembly---where L5 Centromere Assembly drives downstream G2/M Transition and Spindle Checkpoint features ($d = -1.37$ to $-1.76$), recapitulating the known centromere$\to$kinetochore$\to$spindle checkpoint cascade~\citep{musacchio2007spindle,malumbres2009cell} (Supplementary Note~1; Table~\ref{tab:circuits_gf}; Figure~\ref{fig:mitotic})---and neurodevelopment-proteostasis coupling, where an L0 Nervous System Development hub (F146) drives proteasomal, endosomal, and Golgi targets across 7 downstream layers (128--142 shared terms per edge), mirroring the known dependence of neurons on protein quality control systems~\citep{ross2004proteostasis,labbadia2015proteostasis} (Supplementary Note~2; Figure~\ref{fig:neuro_hub}). Metabolic circuits further demonstrate cross-depth persistence: L5 DNA Metabolic Process reinforces itself at L6 ($d = -1.64$) and L10 ($d = -1.77$), while L0 Cholesterol Biosynthesis (out-degree 5,096) acts as a broad metabolic hub~\citep{simons2000cholesterol}.

\begin{table}[t]
\centering
\caption{\textbf{Extended interpretable causal circuits in Geneformer K562/K562.} Circuits are organized by biological theme. All edges are inhibitory (negative $d$). ``Shared'' indicates the number of shared ontology terms between source and target features.}
\label{tab:circuits_gf}
\smallskip
\begin{tabular}{llrrl}
\toprule
Source & Target & $d$ & Shared & Biological Theme \\
\midrule
\multicolumn{5}{l}{\emph{DNA damage response}} \\
L0 DNA Repair & L1 DNA Damage Resp. & $-1.87$ & 113 & Damage detection$\to$response \\
L0 DNA Repair & L6 Kinetochore & $-3.47$ & --- & Damage$\to$mitotic checkpoint \\
L5 DNA Metabolic & L7 G2/M Transition & $-2.45$ & --- & Replication$\to$division \\
L5 DNA Metabolic & L11 Kinetochore & $-2.39$ & --- & Replication$\to$spindle \\
\midrule
\multicolumn{5}{l}{\emph{Mitotic apparatus}} \\
L5 Cell Cycle G2/M & L11 Spindle Micro. & $<-1.4$ & --- & Commitment$\to$assembly \\
L5 Centromere Asm. & L11 G2/M Transition & $-1.37$ & 73 & Centromere$\to$checkpoint \\
L11 Cytokinesis & L15 Spindle Ckpt. & $<-0.9$ & --- & Late mitotic feedback \\
\midrule
\multicolumn{5}{l}{\emph{Neurodevelopment--proteostasis}} \\
L0 Nerv.\ Sys.\ Dev. & L1 Endosome Org. & $-1.32$ & 142 & Neurodegeneration \\
L0 Nerv.\ Sys.\ Dev. & L2 Proteasome Cat. & $-0.96$ & 140 & Protein quality ctl. \\
L0 Nerv.\ Sys.\ Dev. & L6 Protein Catabolism & $-1.27$ & 139 & NF-$\kappa$B signaling \\
\midrule
\multicolumn{5}{l}{\emph{Metabolism}} \\
L5 DNA Metabolic & L6 DNA Metabolic & $-1.64$ & 126 & Self-reinforcing \\
L5 DNA Metabolic & L10 DNA Metabolic & $-1.77$ & 115 & Long-range maint. \\
\bottomrule
\end{tabular}
\end{table}

\begin{figure}[H]
\centering
\includegraphics[width=\textwidth]{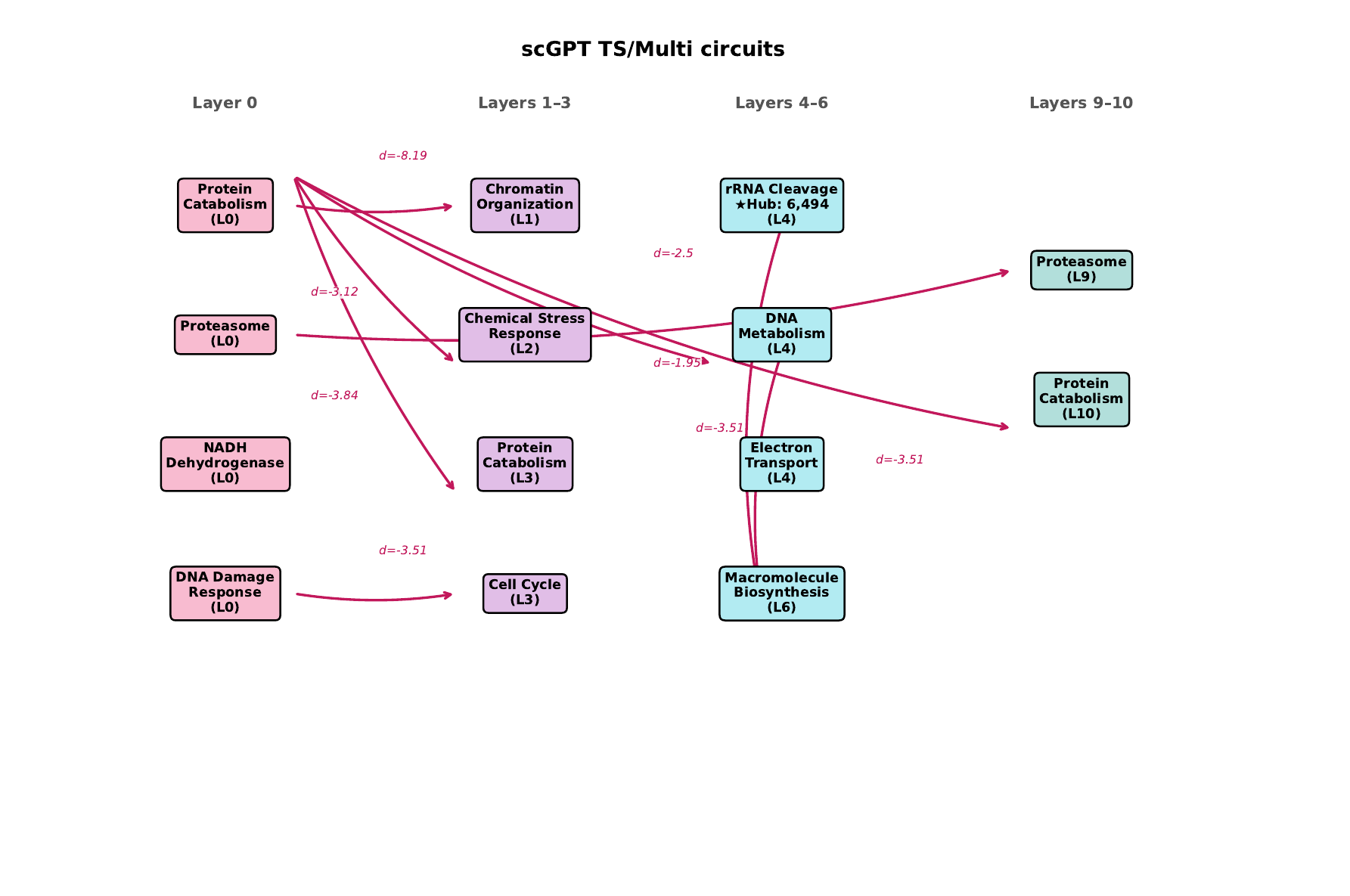}
\caption{\textbf{scGPT TS/Multi biological circuit architecture.} 14 features across layers 0--10. The L0 Protein Catabolism hub drives the strongest individual edges in either model ($d = -8.19$ to Chromatin Organization, $d = -6.10$ to DNA Metabolism). Both models show cross-layer proteasome persistence and DDR$\to$cell cycle connections.}
\label{fig:circuits_scgpt}
\end{figure}

\subsection{Causal effects persist across the full network depth}
\label{sec:attenuation}

\begin{figure}[H]
\centering
\includegraphics[width=\textwidth]{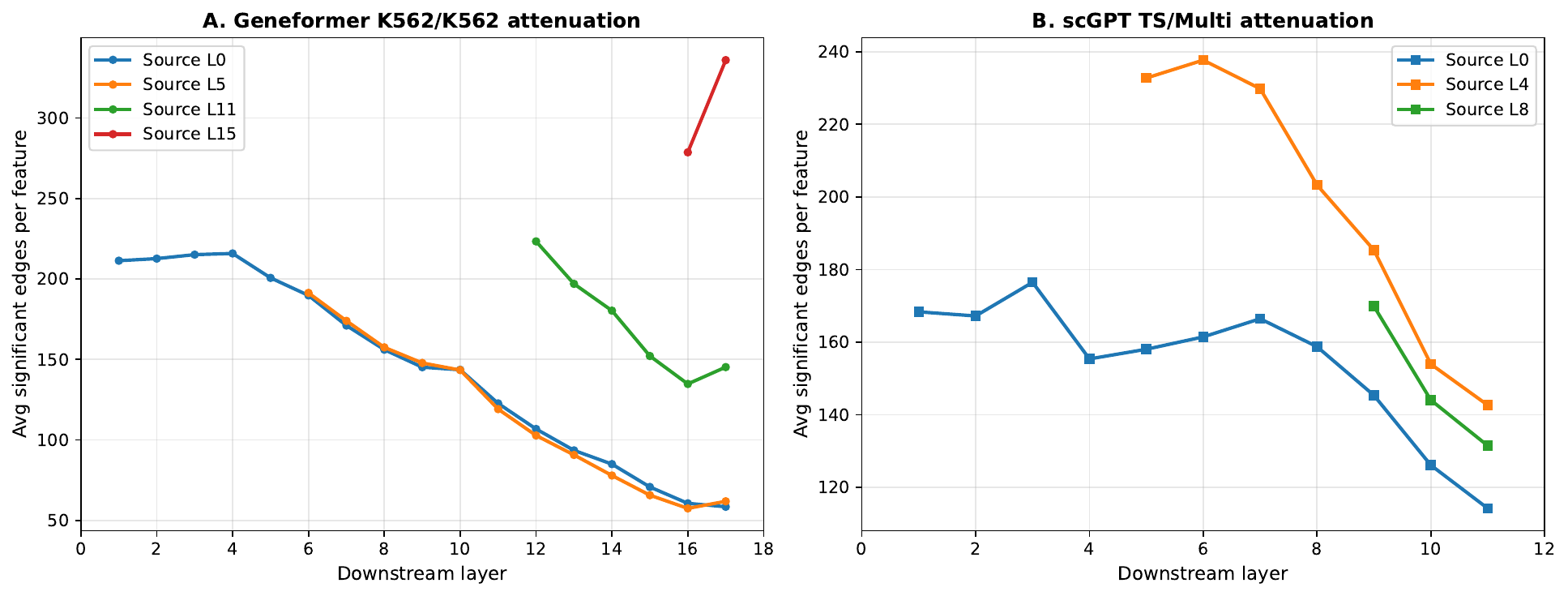}
\caption{\textbf{Causal effect attenuation across network depth.} \textbf{(A)}~Geneformer K562/K562: L0 effects persist across all 17 downstream layers, maintaining $>$200 edges for 5 layers before decaying. L15$\to$L17 effects increase, suggesting late-layer consolidation. \textbf{(B)}~scGPT TS/Multi: L0 effects show a flat plateau through $\sim$L6, then gradual decline. L4 features are the most broadly connected, in contrast to Geneformer where L0 dominates.}
\label{fig:attenuation}
\end{figure}

A fundamental question is how far causal effects propagate: does ablating an L0 feature affect only L1, or do effects persist to L17? Figure~\ref{fig:attenuation}A presents the attenuation curves for Geneformer.

L0 effects are remarkably persistent: they maintain ${\sim}$200--215 significant edges per downstream layer for the first 5 layers (L1--L5), with a slight \emph{increase} at L3 (215) before decaying. Even at L17---17 layers downstream---L0 features still produce 58 significant edges per feature. L5 effects decay more linearly, dropping from 191 to 62 over 12 layers. L11 effects remain strong over their 6 downstream layers (223$\to$145). L15$\to$L17 effects actually \emph{increase} from 279 to 336, suggesting strong local coupling at late layers that may reflect the model consolidating information for output prediction.

This persistence has important implications: early-layer features contain foundational biological information that the entire network depends on. The model does not re-derive this information at each layer; rather, it propagates and transforms it through the depth.

\subsection{Statistical co-activation validates causal edges}
\label{sec:pmi}

We compared causal circuit edges with the PMI-based statistical co-activation graph computed independently in Phase~2~\citep{kendiukhov2025sae_atlas}. Source features were selected by different criteria (annotation quality for causal tracing vs.\ co-activation frequency for PMI), resulting in zero source feature overlap. However, \textbf{target feature overlap is 91--95\%} (Table~\ref{tab:pmi}): the downstream features identified as causally influenced by circuit tracing are nearly the same features identified as statistically co-activated by PMI.

This convergence validates both methods. Statistical co-activation captures genuine information flow (not noise), while causal tracing additionally reveals directionality, effect magnitude, and sign (excitatory vs.\ inhibitory). The two approaches are complementary views of the same underlying computational structure.

\begin{table}[t]
\centering
\caption{\textbf{PMI and causal edge comparison (Geneformer K562/K562).} Target overlap measures the fraction of causal targets that also appear as PMI targets at the same layer pair.}
\label{tab:pmi}
\smallskip
\begin{tabular}{lrrr}
\toprule
Layer Pair & PMI Targets & Causal Targets & Target Overlap \\
\midrule
L0$\to$L5 & 4,101 & 1,113 & 90.6\% \\
L5$\to$L11 & 4,369 & 996 & 94.8\% \\
L11$\to$L17 & 4,205 & 881 & 91.7\% \\
\bottomrule
\end{tabular}
\end{table}

\subsection{Multi-tissue SAEs produce more biologically coherent circuits}
\label{sec:multitissue}

To test whether SAE training data affects circuit properties, we repeated circuit tracing using multi-tissue SAEs (trained on K562 + Tabula Sapiens cells; available at layers 0, 5, 11, 17 only) with the same 200 K562 cells. This yields the K562/Multi condition (Table~\ref{tab:aggregate}).

\begin{figure}[H]
\centering
\includegraphics[width=\textwidth]{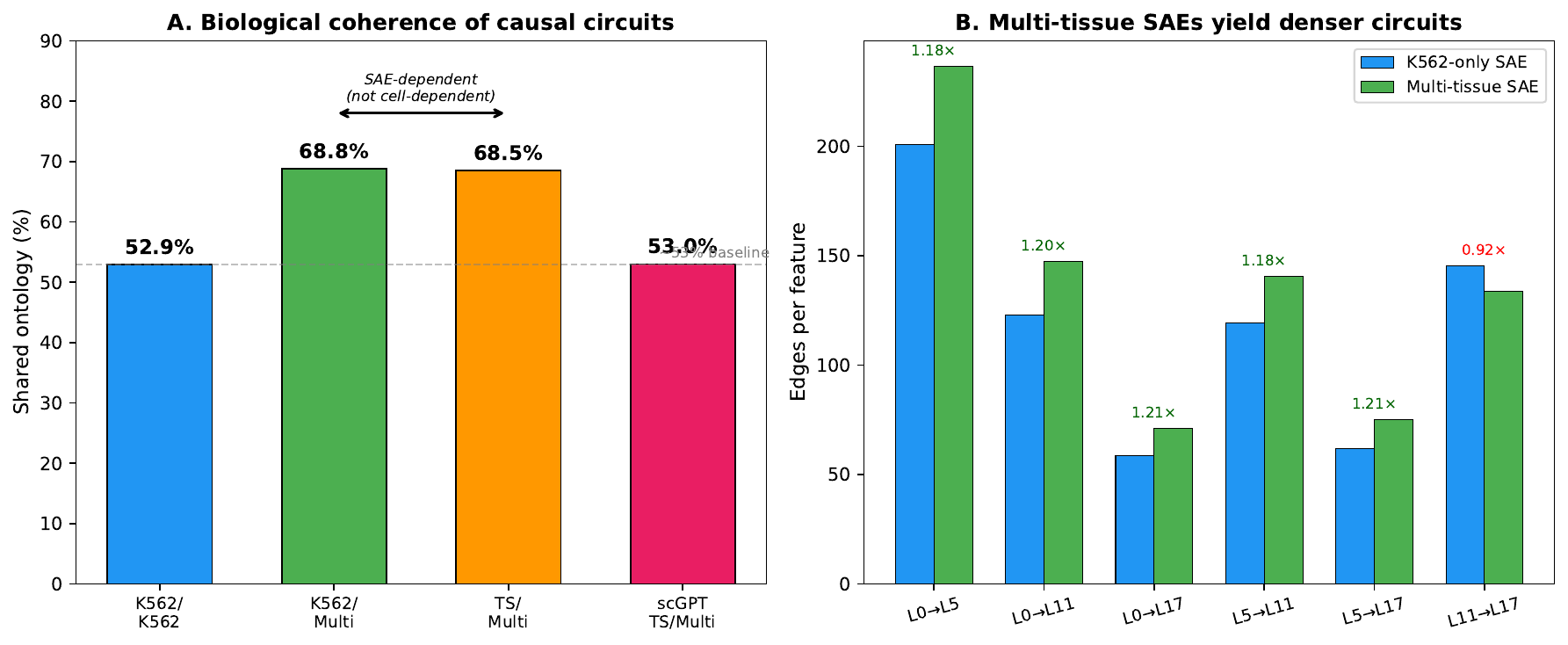}
\caption{\textbf{Biological coherence of causal circuits.} \textbf{(A)}~Shared ontology fraction across conditions. The $\sim$53\% baseline (dashed line) is invariant to model architecture and cell type. Multi-tissue SAEs achieve 68.8\%, and critically, this value is identical whether processing K562 or Tabula Sapiens cells (68.8\% vs.\ 68.5\%), demonstrating SAE-dependence, not cell-dependence. \textbf{(B)}~Per-layer-pair edge density: multi-tissue SAEs yield 18--21\% more edges per feature at most layer pairs (green ratios), suggesting broader feature connectivity from diverse training.}
\label{fig:coherence}
\end{figure}

Multi-tissue circuits show two notable differences from K562-only circuits. First, multi-tissue SAE features are \textbf{18--21\% more broadly connected}: at matched layer pairs, each feature produces 18--21\% more significant downstream edges (Figure~\ref{fig:coherence}B). The exception is L11$\to$L17, where K562-only edges are slightly denser (ratio 0.92). Second, and more importantly, biological coherence increases substantially: \textbf{68.8\% of multi-tissue edges share ontology terms} between source and target, compared to 52.9\% for K562-only ($+$16 percentage points). This suggests that features trained on diverse cell types capture cleaner, more universal biological programs that form tighter circuits.

The inhibitory fraction is virtually identical (79.9\% vs.\ 80.1\%), indicating that the fundamental computational architecture---predominantly inhibitory information flow---is invariant to SAE training data.

Top hub features in multi-tissue circuits differ from K562-only: the dominant hubs are Histone Modification (L0, 2,146 targets), RNA Processing (L0, 1,273 targets), and DNA Damage Response (L0, 562 targets). The multi-tissue DDR cascade (Figure~\ref{fig:cascade}) represents one of the most biologically compelling findings: a complete pathway from damage sensing through checkpoint activation to cell cycle arrest, encoded as a directed causal circuit spanning four network layers.

\subsection{Biological coherence depends on the SAE lens, not the input cells}
\label{sec:ts_cells}

To disentangle SAE effects from cell-type effects, we repeated circuit tracing with the same multi-tissue SAEs but using 200 Tabula Sapiens cells (67 immune, 67 kidney, 66 lung; stratified sampling across 88 cell types) instead of K562 cells. This yields the TS/Multi condition.

\begin{figure}[H]
\centering
\includegraphics[width=\textwidth]{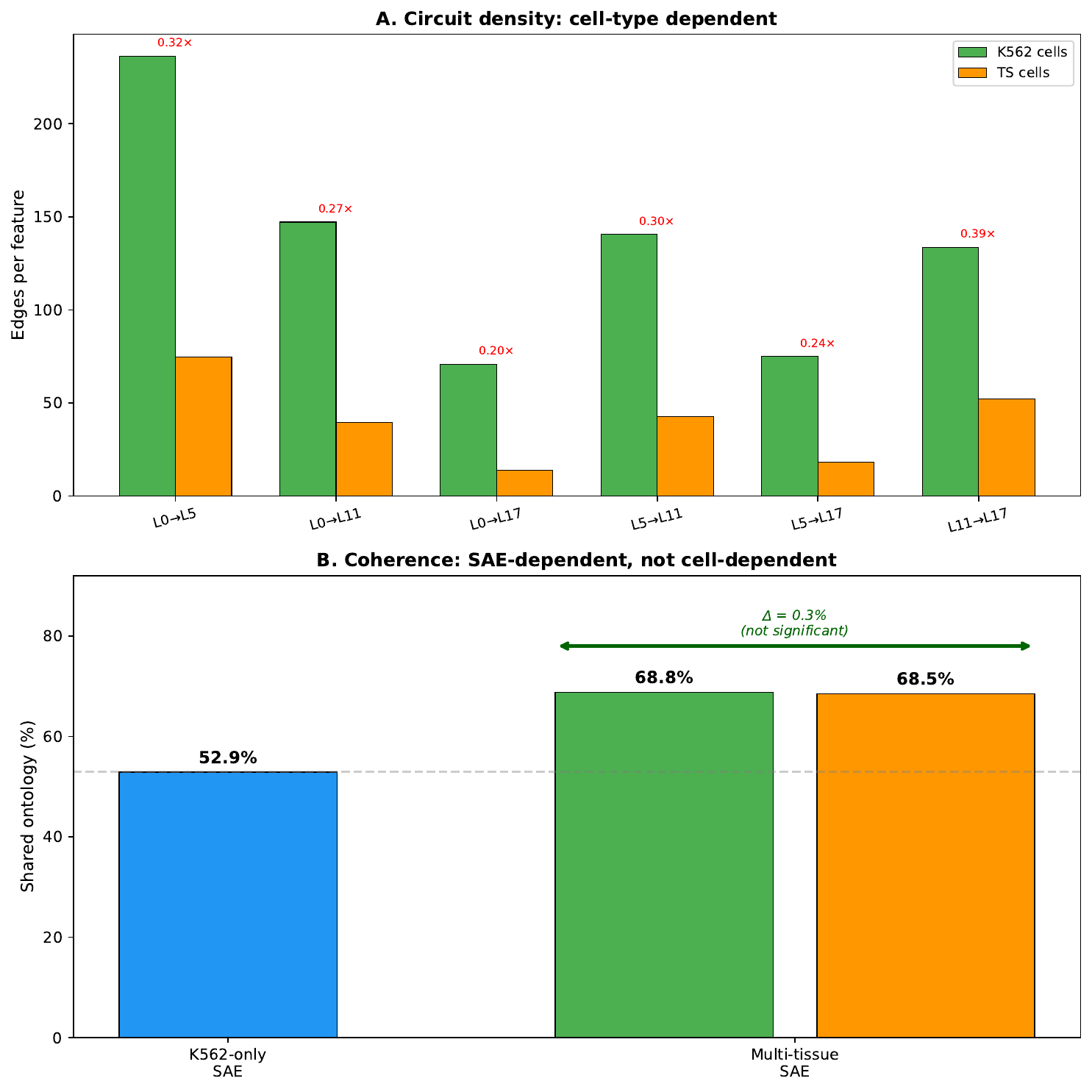}
\caption{\textbf{Dissociating SAE effects from cell-type effects.} \textbf{(A)}~Circuit density is strongly cell-type-dependent: K562 cells produce 3--5$\times$ more edges per feature than Tabula Sapiens cells through the same multi-tissue SAEs (ratios in red). \textbf{(B)}~Biological coherence is SAE-dependent, not cell-dependent: multi-tissue SAEs achieve 68.8\% (K562 cells) vs.\ 68.5\% (TS cells)---a difference of only 0.3 percentage points (green bracket).}
\label{fig:celltype}
\end{figure}

The results reveal a clean dissociation (Figure~\ref{fig:celltype}). Tabula Sapiens circuits are \textbf{3--5$\times$ sparser}: 75 edges per feature at L0$\to$L5 vs.\ 236 for K562 cells with the same SAEs (ratio 0.32; Figure~\ref{fig:celltype}A). Effect sizes are substantially weaker (mean $|d| = 0.72$ vs.\ 0.98; only 10.4\% with $|d| > 1.0$ vs.\ 34.4\%; Figure~\ref{fig:effect_sizes}B). This sparsity likely reflects cell-type mismatch: SAE features were selected by annotation quality derived from K562 training data, making them less salient for non-K562 cell types.

However, \textbf{biological coherence is unchanged}: 68.5\% of TS/Multi edges share ontology terms, virtually identical to the 68.8\% observed with K562/Multi (Figure~\ref{fig:celltype}B). This demonstrates that biological coherence is a property of the SAE feature dictionary---the lens through which we view the model---not the input data being processed.

Specific circuits are preserved but attenuated. The DDR cascade shows $d = -3.84$ (L0$\to$L5 DDR) with K562 cells but only $d = -1.30$ with TS cells; Centromere Assembly$\to$G2/M Transition drops from $-1.76$ to $-0.87$. Interestingly, one strong circuit in TS cells involves NADH Dehydrogenase: L0 NADH Dehydrogenase $\to$ L5 NADH Dehydrogenase ($d = -2.17$), reflecting the universal importance of mitochondrial energy metabolism across all tissue types. A G1/S Transition feature drives a Centromere Assembly feature ($d = -3.08$), capturing the biological link between DNA replication licensing and centromere maturation.

TS/Multi circuits are \textbf{89.4\% inhibitory}---the highest of any condition---compared to ${\sim}80\%$ for both K562 conditions. When processing unfamiliar cell types, each feature becomes even more ``necessary''---there is less redundancy in representing non-training-domain biology, so ablation more consistently impairs downstream computation.

\subsection{scGPT circuit tracing reveals a fundamentally different computational architecture}
\label{sec:scgpt}

We applied identical circuit tracing methodology to scGPT whole-human (12 layers, $d{=}512$, 2,048 features per layer), using 200 Tabula Sapiens cells with scGPT's natively multi-tissue SAEs (trained on 3,000 TS cells across immune, kidney, and lung tissues). Source features (30 per layer at L0, L4, L8) were selected by annotation quality from the scGPT feature atlas. All 12 layers have SAEs, enabling comprehensive downstream tracing.

The scGPT circuit graph contains \textbf{31,380 significant causal edges} connecting 90 source features to 1,960 unique target features---95.7\% of all 2,048 features per layer (Tables~\ref{tab:aggregate} and~\ref{tab:scgpt_perlayer}). The total computation required only 37.2 minutes on Apple Silicon, reflecting scGPT's smaller dimensionality.

\begin{table}[t]
\centering
\caption{\textbf{Per-source-layer circuit statistics for scGPT.} All 12 layers have SAEs, so downstream tracing is comprehensive.}
\label{tab:scgpt_perlayer}
\smallskip
\begin{tabular}{lrrrrr}
\toprule
Source & Downstream & Passes & Time & Total Edges & Avg/Feature \\
\midrule
L0 & 11 (L1--L11) & 6,194 & 17.0 min & 49,777 & 1,659 \\
L4 & 7 (L5--L11) & 6,187 & 14.2 min & 41,409 & 1,380 \\
L8 & 3 (L9--L11) & 6,114 & 5.9 min & 11,361 & 379 \\
\bottomrule
\end{tabular}
\end{table}

\subsubsection{Stronger individual effects, more balanced dynamics}

scGPT edges are substantially stronger: mean $|d| = 1.40$ vs.\ 1.05 for Geneformer K562/K562. 65.2\% of scGPT edges exceed $|d| > 1.0$, compared to 41.4\% for Geneformer (Figure~\ref{fig:effect_sizes}B). This likely reflects dimensionality: with 2,048 features (vs.\ 4,608), each scGPT feature carries a larger fraction of the total representation.

scGPT circuits show a more balanced \textbf{inhibitory/excitatory ratio}: 65.5\% inhibitory vs.\ 80.1\% for Geneformer (Figure~\ref{fig:effect_sizes}C). The 34.5\% excitatory fraction indicates more competitive dynamics: features more frequently suppress each other, so removing one releases others from inhibition. Geneformer's 80/20 ratio suggests a more cooperative, dependency-based architecture.

\subsubsection{Biological coherence converges across architectures}

Despite fundamentally different architectures, training objectives, and training data, scGPT achieves \textbf{53.0\% shared ontology}---virtually identical to Geneformer K562/K562 at 52.9\% (Figure~\ref{fig:coherence}A). This convergence at ${\sim}53\%$ across two independent models strongly suggests that this value reflects a property of biological knowledge organization rather than any model-specific computation.

\begin{figure}[H]
\centering
\includegraphics[width=\textwidth]{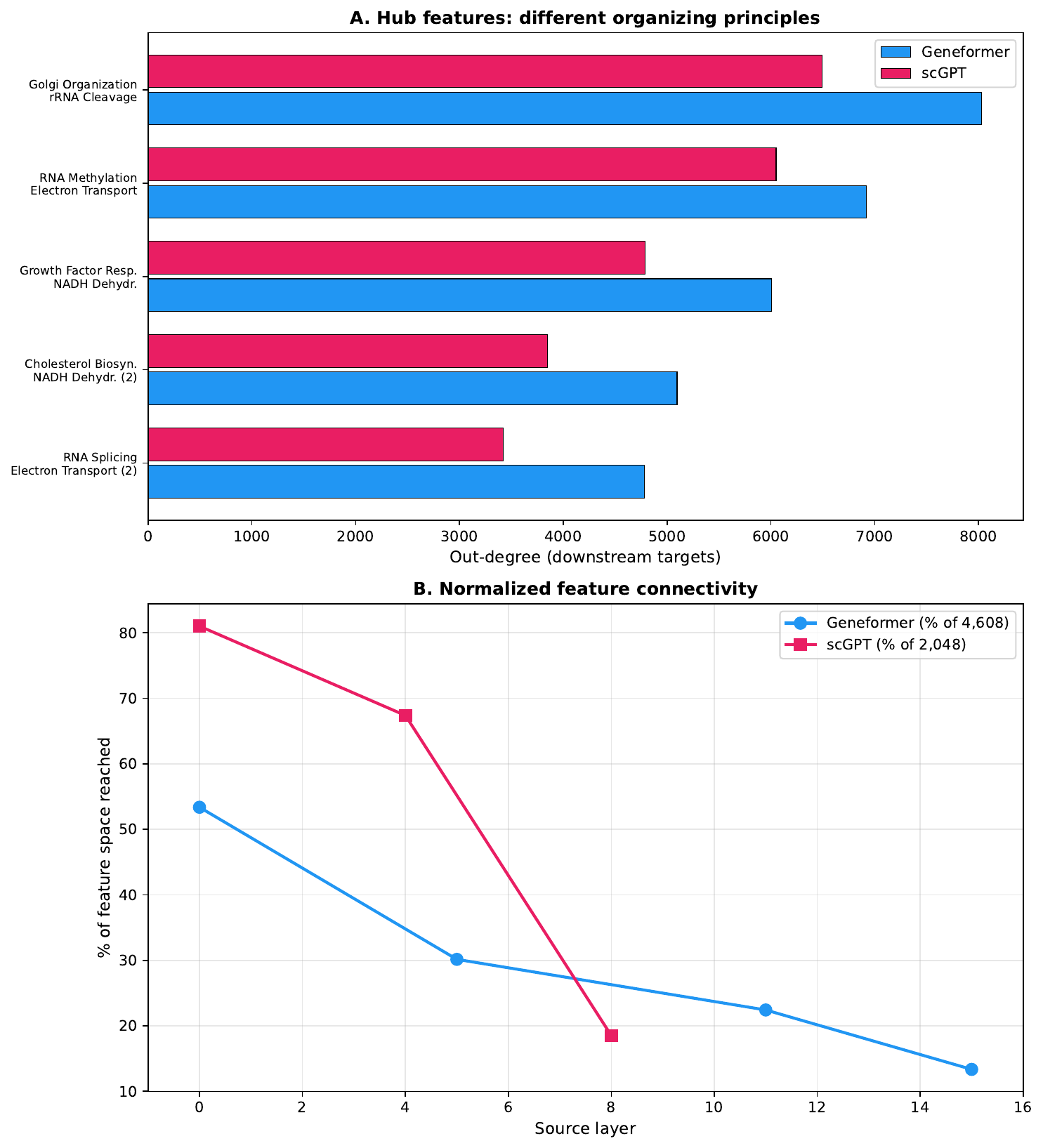}
\caption{\textbf{Cross-model comparison of circuit architecture.} \textbf{(A)}~Hub features reveal different organizing principles: Geneformer (blue) hubs center on Golgi organization, RNA processing, and growth factor response; scGPT (pink) hubs center on rRNA processing and mitochondrial electron transport. \textbf{(B)}~Normalized connectivity: the percentage of the downstream feature space reached per source feature. scGPT features reach a larger fraction of the available feature space (up to 81\% at L0), reflecting fewer total features but broader individual influence.}
\label{fig:crossmodel}
\end{figure}

\subsubsection{Energy metabolism as organizing hub}

scGPT's hub features differ qualitatively from Geneformer's (Table~\ref{tab:hubs_scgpt}; Figure~\ref{fig:crossmodel}A). The top hubs are dominated by mitochondrial electron transport: NADH Dehydrogenase Complex Assembly (L0, 4,785 and 3,849 targets) and Aerobic Electron Transport Chain (L0, 3,420 targets; L4, 6,050 targets). scGPT has organized its internal representation around \emph{energy metabolism} as a central axis---biologically sensible given that energy status is a fundamental cellular variable influencing nearly all other processes (Supplementary Note~4). At L4, Endonucleolytic Cleavage for rRNA processing reaches an out-degree of 6,494---the single highest-connectivity feature across either model (Supplementary Note~6).

\begin{table}[t]
\centering
\caption{\textbf{Top hub features in scGPT circuits.} Mitochondrial electron transport features dominate, in contrast to Geneformer's chromatin and RNA processing hubs.}
\label{tab:hubs_scgpt}
\smallskip
\begin{tabular}{llr}
\toprule
Feature & Biology & Out-Degree \\
\midrule
L4\_F446 & Endonucleolytic Cleavage (rRNA) & 6,494 \\
L4\_F1643 & Aerobic Electron Transport Chain & 6,050 \\
L0\_F552 & NADH Dehydrogenase Complex Assembly & 4,785 \\
L0\_F590 & NADH Dehydrogenase Complex Assembly & 3,849 \\
L0\_F233 & Aerobic Electron Transport Chain & 3,420 \\
L0\_F880 & Golgi Organization & 3,133 \\
L4\_F725 & Heart Development & 3,046 \\
\bottomrule
\end{tabular}
\end{table}

\subsubsection{Interpretable stress response and protein quality control circuits}

\begin{figure}[H]
\centering
\includegraphics[width=\textwidth]{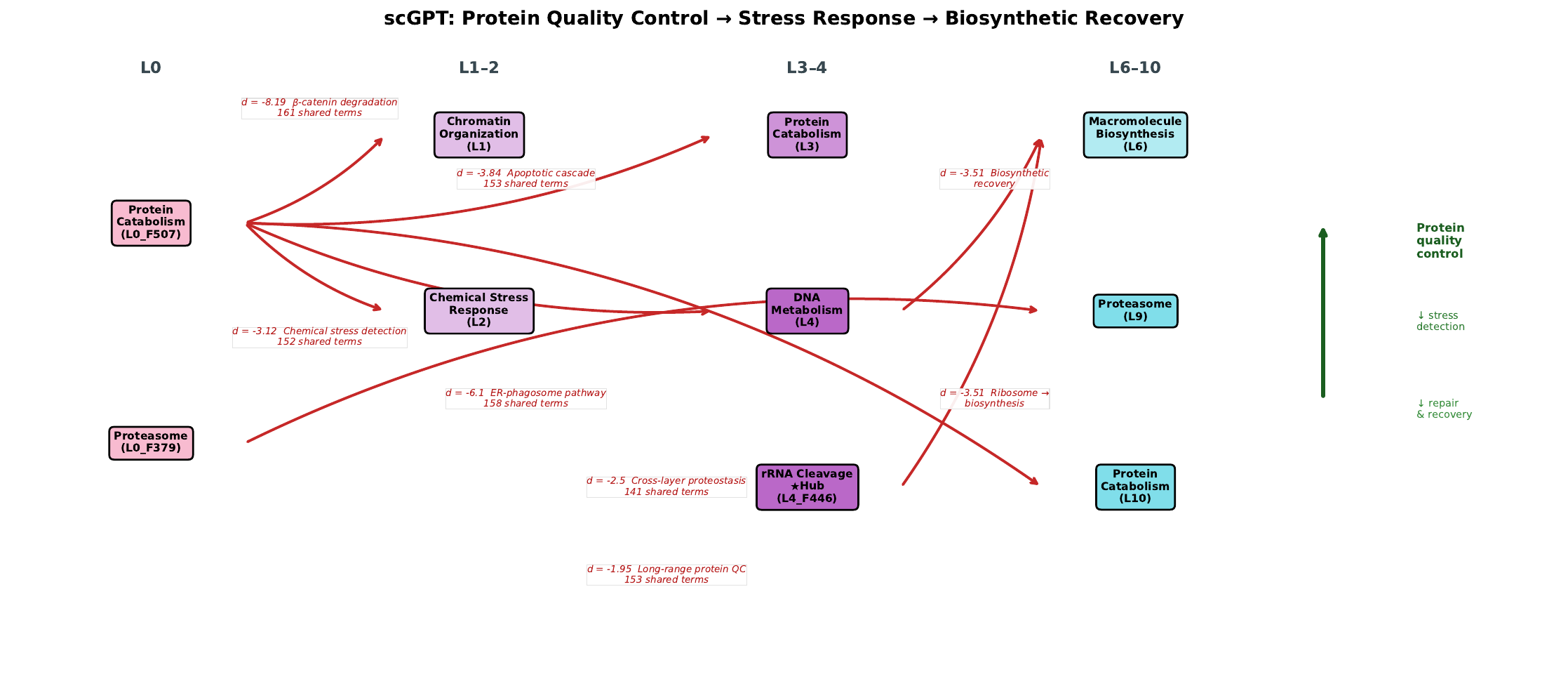}
\caption{\textbf{scGPT proteostasis and stress response cascade.} The L0 Protein Catabolism hub (F507) radiates causal influence across multiple downstream pathways. The strongest edge ($d = -8.19$) connects to L1 Chromatin Organization via beta-catenin degradation~\citep{nusse2017wnt}, sharing 161 ontology terms. A coherent biological progression flows from protein quality control (L0) $\to$ stress detection (L2) $\to$ DNA repair (L4) $\to$ biosynthetic recovery (L6), paralleling the integrated stress response~\citep{walter2011unfolded}. The L0 Proteasome feature independently drives L9 Proteasome ($d = -2.50$, 141 shared terms), demonstrating cross-layer persistence of proteostasis monitoring~\citep{labbadia2015proteostasis}. The L4 rRNA Cleavage hub (out-degree 6,494) provides an alternative route to macromolecule biosynthesis via ribosome biogenesis~\citep{boulon2010nucleolus}.}
\label{fig:scgpt_cascade}
\end{figure}

The strongest individual scGPT circuits reveal a coherent stress response architecture (Figure~\ref{fig:circuits_scgpt}; Table~\ref{tab:circuits_scgpt}; Figure~\ref{fig:scgpt_cascade}). The strongest edge---L0 Protein Catabolism $\to$ L1 Chromatin Organization ($d = -8.19$, 161 shared terms)---is linked through stress response and beta-catenin degradation pathways. Beta-catenin is a dual-function protein: in the Wnt pathway it acts as a transcriptional coactivator, but it is constitutively degraded by the proteasome when Wnt signals are absent~\citep{nusse2017wnt}. The model has learned that protein catabolism programs are causally upstream of chromatin organization---reflecting the biological reality that proteasomal degradation of transcription factors directly reshapes the chromatin landscape~\citep{labbadia2015proteostasis}.

A multi-layer cascade emerges from this hub: L0 Protein Catabolism drives L2 Chemical Stress Response ($d = -3.12$, 152 terms), L3 Protein Catabolism ($d = -3.84$, 153 terms: apoptosis), L4 DNA Metabolism ($d = -6.10$, 158 terms: ER-phagosome), and ultimately L10 Protein Catabolism ($d = -1.95$, 153 terms). The L4 DNA Metabolism node then drives L6 Macromolecule Biosynthesis ($d = -3.51$, 153 terms). This cascade represents the biological progression: \emph{protein quality control $\to$ stress detection $\to$ DNA repair $\to$ biosynthetic recovery}---a coherent trajectory through the unfolded protein response~\citep{hetz2012unfolded} and integrated stress response~\citep{walter2011unfolded} pathways.

A second prominent circuit family involves the ubiquitin-proteasome system. L0 Proteasome (F379) drives downstream Proteasome features at nearly every subsequent layer (16 of the top 50 circuits), with the strongest connection to L9 Proteasome ($d = -2.50$, 141 shared terms). This cross-layer persistence of proteasome features suggests that protein degradation status is tracked throughout the model's computation, consistent with the biological importance of proteostasis as a cell-wide quality control system~\citep{labbadia2015proteostasis}. Full scGPT stress response and proteasome circuit descriptions are provided in Supplementary Note~3.

\begin{table}[t]
\centering
\caption{\textbf{Extended interpretable causal circuits in scGPT.} Circuits organized by biological theme.}
\label{tab:circuits_scgpt}
\smallskip
\begin{tabular}{llrrl}
\toprule
Source & Target & $d$ & Shared & Biological Theme \\
\midrule
\multicolumn{5}{l}{\emph{Stress response cascade}} \\
L0 Protein Catabolism & L1 Chromatin Org. & $-8.19$ & 161 & $\beta$-catenin degrad. \\
L0 Protein Catabolism & L2 Chemical Stress & $-3.12$ & 152 & Stress detection \\
L0 Protein Catabolism & L4 DNA Metabolism & $-6.10$ & 158 & ER-phagosome \\
L4 DNA Metabolism & L6 Macromol.\ Biosyn. & $-3.51$ & 153 & Biosynthetic recovery \\
\midrule
\multicolumn{5}{l}{\emph{Ubiquitin-proteasome persistence}} \\
L0 Proteasome & L9 Proteasome & $-2.50$ & 141 & Cross-layer QC \\
L0 Protein Catabolism & L3 Protein Catab. & $-3.84$ & 153 & Apoptotic cascade \\
L0 Protein Catabolism & L10 Protein Catab. & $-1.95$ & 153 & Long-range maint. \\
\midrule
\multicolumn{5}{l}{\emph{DDR and cell cycle}} \\
L0 DDR & L3 Cell Cycle & $-3.51$ & --- & Damage$\to$arrest \\
L4 rRNA Cleavage & L6 Macromol.\ Biosyn. & $-3.51$ & --- & Ribosome$\to$biosyn. \\
\bottomrule
\end{tabular}
\end{table}

\subsubsection{Mid-layer features are the most connected}

scGPT shows a distinct attenuation pattern compared to Geneformer (Figure~\ref{fig:attenuation}B). L0 effects show a remarkable flat plateau---maintaining $\sim$155--176 edges per layer through L6 with no decay---before declining to 114 at L11. L4 features show the densest connectivity at 197.9 significant edges per downstream layer, exceeding both L0 (154.3) and L8 (148.5). This contrasts with Geneformer, where L0 features are the most broadly connected. The scGPT pattern suggests a ``funnel then broadcast'' architecture, where mid-network features serve as integration hubs.

\subsection{Circuits that span multiple biological domains suggest cross-program coordination}
\label{sec:cross_domain}

Among the 47\% of causal edges that do \emph{not} share ontology terms, some connect features from distinct biological domains, suggesting that the models have learned cross-program coordination that transcends traditional pathway boundaries.

In Geneformer, L0 MAPK Cascade features (F2829) causally influence features across multiple downstream domains including protein catabolism, vesicle transport, and RNA processing. MAPK signaling is indeed a master regulator that coordinates diverse cellular responses---from proliferation to apoptosis to differentiation~\citep{dhillon2007map}---so the model's representation of MAPK as a cross-domain hub is biologically justified even though the downstream targets may not share MAPK-specific ontology terms.

In scGPT, the rRNA Endonucleolytic Cleavage hub (L4, out-degree 6,494) drives features in macromolecule biosynthesis, protein catabolism, and DNA metabolism. Ribosomal RNA processing is a rate-limiting step for protein synthesis; its disruption triggers nucleolar stress, p53 activation, and cell cycle arrest~\citep{boulon2010nucleolus}---connecting ribosome biogenesis to virtually all other cellular programs. The model has learned this connectivity, encoding rRNA processing as a computational hub that broadcasts information to diverse downstream pathways.

These cross-domain circuits may represent the most biologically interesting edges for hypothesis generation, as they capture functional relationships that cross traditional pathway boundaries and may not be well-represented in existing ontology databases.

\subsection{Systematic biological knowledge extraction across 96,892 edges}
\label{sec:systematic}

The preceding sections characterized ${\sim}200$ circuits through manual examination. To systematically extract the full biological knowledge, we annotated all 96,892 causal edges across all four conditions with source and target biological domain labels (GO Biological Process terms from feature annotations), yielding 37,088 edges (38.3\%) where both endpoints are annotated. This enables quantitative analysis of the complete circuit landscape (Table~\ref{tab:knowledge}).

\subsubsection{Cross-model consensus reveals 1,142 conserved biological circuits}

\begin{figure}[H]
\centering
\includegraphics[width=\textwidth]{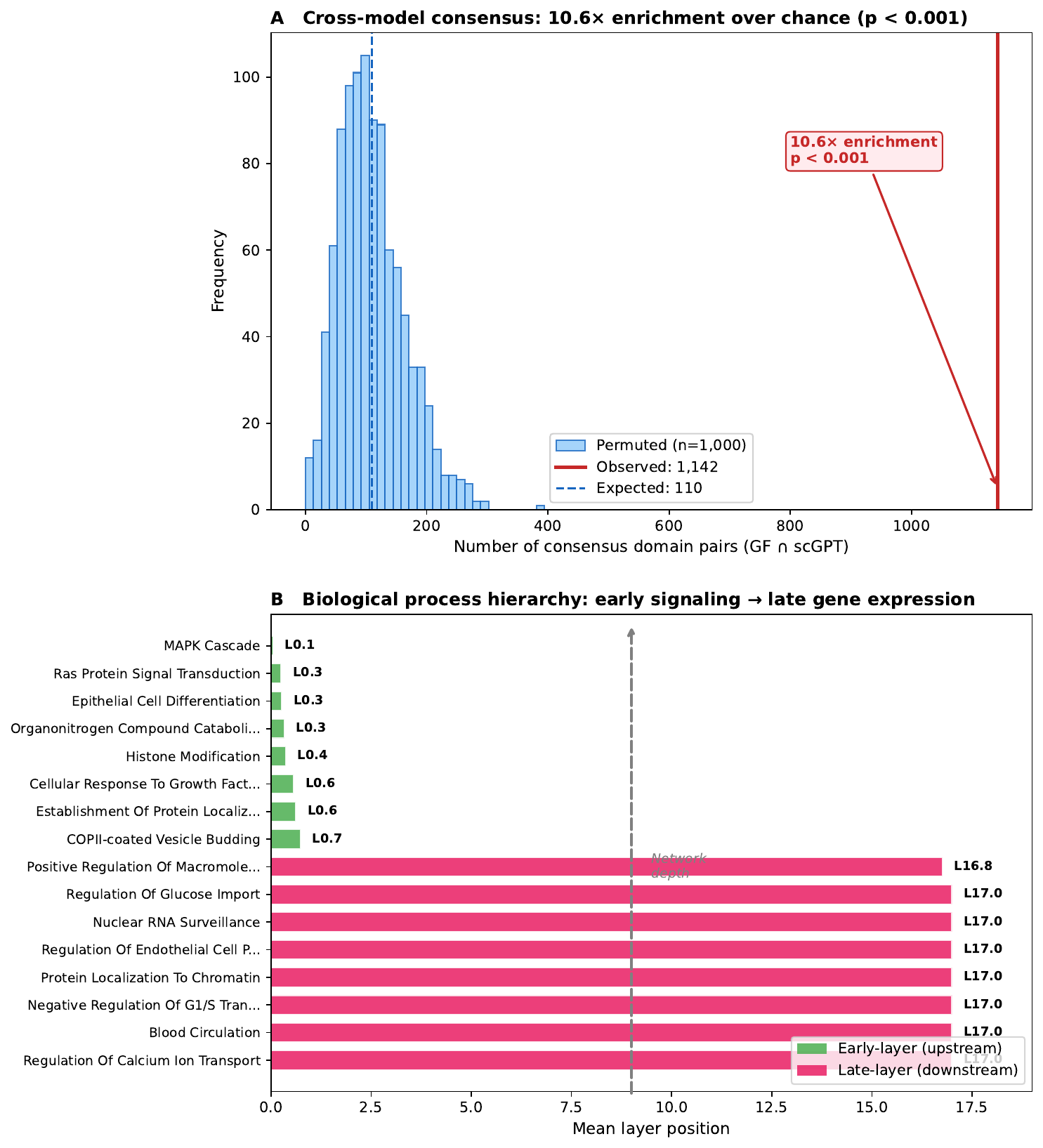}
\caption{\textbf{Systematic biological knowledge extraction.} \textbf{(A)}~Permutation test for cross-model consensus. 1,142 domain pairs are conserved between Geneformer and scGPT, 10.6$\times$ more than expected by random label permutation ($p < 0.001$, 1,000 permutations). \textbf{(B)}~Biological process hierarchy derived from layer position. Early-layer processes (green) include signaling cascades (MAPK, Ras) and cellular structure (COPII vesicles, protein localization); late-layer processes (pink) include gene expression regulation, RNA surveillance, and cell fate decisions. This temporal ordering matches known biology: signaling precedes transcriptional output.}
\label{fig:knowledge}
\end{figure}

We identified domain pairs (source biological process $\to$ target biological process) present in at least one Geneformer condition AND in scGPT. Of 13,698 unique Geneformer domain pairs and 3,511 scGPT pairs, \textbf{1,142 pairs appear in both models}---10.6$\times$ more than expected by chance (permutation test, 1,000 permutations: expected 107.3, $p < 0.001$; Figure~\ref{fig:knowledge}A). Of these, 303 are \textbf{high-confidence} pairs where both models show mean $|d| > 1.0$.

Top consensus circuits include: Golgi Organization $\to$ Protein Insertion Into Membrane (GF $|d|{=}4.75$, scGPT $|d|{=}5.23$), Cholesterol Biosynthesis $\to$ Sterol Biosynthesis (GF $|d|{=}4.44$, scGPT $|d|{=}1.54$), and Maturation of SSU-rRNA $\to$ RNA Methylation (GF $|d|{=}1.67$, scGPT $|d|{=}5.00$). The strong enrichment demonstrates that both models, despite different architectures, training data, and objectives, have converged on a shared biological circuit structure that reflects genuine functional relationships rather than model-specific artifacts (Supplementary Note~7).

\begin{table}[H]
\centering
\caption{\textbf{Systematic biological knowledge extraction summary.} Results from automated analysis of all 96,892 causal edges across four conditions.}
\label{tab:knowledge}
\smallskip
\begin{tabular}{lrl}
\toprule
Metric & Value & Significance \\
\midrule
Total causal edges & 96,892 & Across 4 conditions \\
Both-annotated edges & 37,088 & 38.3\% annotation rate \\
Unique domain pairs & 16,067 & 1,126 unique domains \\
\midrule
\multicolumn{3}{l}{\emph{Cross-model consensus}} \\
Consensus pairs (GF $\cap$ scGPT) & 1,142 & 10.6$\times$ enrichment \\
High-confidence ($|d|{>}1.0$ both) & 303 & $p < 0.001$ (perm.) \\
\midrule
\multicolumn{3}{l}{\emph{Novel relationships}} \\
Known domain links ($\geq$3 shared genes) & 14,021 & Reference graph \\
Novel edges (not in known biology) & 29,864 & 80.5\% of annotated \\
Novel domain pairs in all 4 conditions & 87 & Top $|d|$ up to 7.21 \\
\midrule
\multicolumn{3}{l}{\emph{Process hierarchy}} \\
Meta-graph nodes/edges & 1,126 / 16,002 & Directed domain graph \\
Feedback loops & 499 & Reciprocal A $\leftrightarrow$ B \\
DDR $\to$ cell cycle edges & 300 & All $\Delta$L $>$ 0 \\
\midrule
\multicolumn{3}{l}{\emph{Tissue specificity}} \\
Immune circuit enrichment & OR = 3.18 & $p < 0.001$ (Fisher's) \\
Immune-specific circuits & 201 & 89\% absent from K562 \\
\bottomrule
\end{tabular}
\end{table}

\subsubsection{Novel relationship discovery identifies 29,864 candidate edges}

We built a reference graph of ``known biology'' by connecting GO BP terms that share $\geq$3 genes (14,021 known links from GO, KEGG, and Reactome databases). Of the 37,088 annotated causal edges, \textbf{29,864 (80.5\%)} connect domain pairs \emph{not} linked in this reference graph, representing candidate novel relationships.

The top novel relationships present in all four conditions include: NADH Dehydrogenase Assembly $\to$ Protein Transport ($|d|{=}7.21$, 20 shared genes), Golgi Organization $\to$ ER Stress Response ($|d|{=}6.29$, 43 shared genes), and Aerobic ETC $\to$ Mitochondrial Translation ($|d|{=}4.67$, 22 shared genes). Many involve cross-compartment functional coupling---mitochondrial processes driving cytoplasmic responses, or secretory pathway organization influencing stress responses---suggesting the models have learned functional relationships that transcend traditional pathway boundaries.

\subsubsection{Biological process hierarchy validates temporal ordering}

The directed meta-graph (1,126 domain nodes, 16,002 edges) reveals a clear temporal ordering of biological processes across network depth (Figure~\ref{fig:knowledge}B). \textbf{Early-layer processes} (mean layer $<$ 1): MAPK cascade (L0.1), Ras signaling (L0.3), histone modification (L0.4), COPII vesicle budding (L0.7). \textbf{Late-layer processes} (mean layer $>$ 16): gene expression regulation, RNA surveillance, protein localization to chromatin. This ordering---signaling $\to$ chromatin modification $\to$ transcriptional output---matches the known biological sequence from signal transduction to gene expression.

Critically, \textbf{300 DDR $\to$ cell cycle edges} were identified, all showing positive layer deltas ($\Delta$L $>$ 0): DNA Repair features appear at earlier layers than their cell cycle targets (e.g., DNA Repair $\to$ Mitotic Chromatid Segregation: $|d|{=}1.27$, $\Delta$L${=}+7.5$; DNA Repair $\to$ Mitotic Cell Cycle Regulation: $|d|{=}1.02$, $\Delta$L${=}+5.9$). This confirms that the model's layer hierarchy faithfully encodes the temporal ordering DNA damage detection $\to$ checkpoint activation $\to$ cell cycle arrest.

\subsubsection{Tissue-specific circuits are enriched in multi-tissue conditions}

\begin{figure}[H]
\centering
\includegraphics[width=\textwidth]{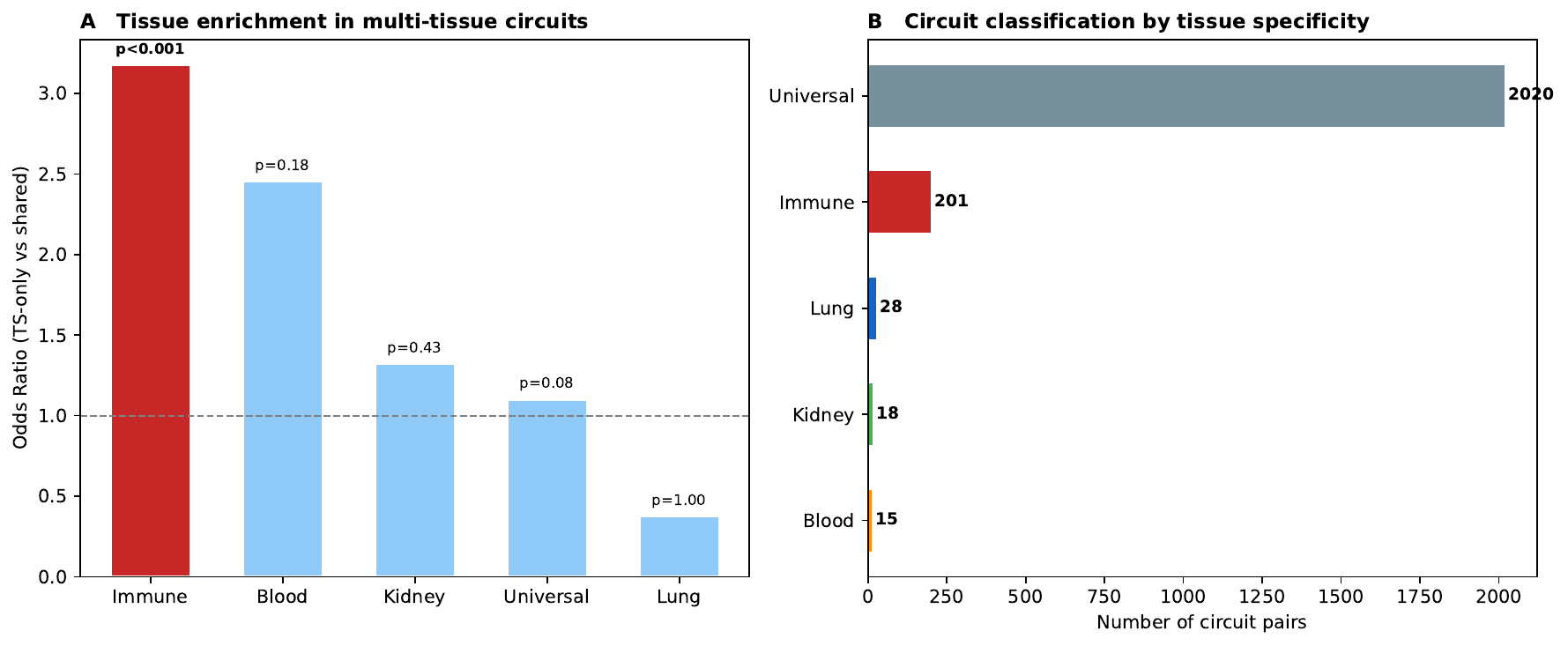}
\caption{\textbf{Tissue-specific circuit activation.} \textbf{(A)}~Enrichment of tissue-related circuits in multi-tissue-specific domain pairs (TS-only) versus shared pairs (present in both K562 and multi-tissue conditions). Immune circuits are significantly enriched (OR${=}3.18$, $p < 0.001$). \textbf{(B)}~Classification of multi-tissue circuit pairs by tissue specificity. Universal circuits (2,020) involve housekeeping processes; immune-specific circuits (201) are predominantly absent from K562.}
\label{fig:tissue_circuits}
\end{figure}

Comparing domain pairs unique to multi-tissue conditions (3,541 pairs) versus those shared with K562 (1,334 pairs) reveals \textbf{significant immune circuit enrichment}: immune-related domain pairs are 3.18$\times$ more frequent in the multi-tissue-specific set (Fisher's exact $p < 0.001$; Figure~\ref{fig:tissue_circuits}A). Of 201 immune-specific circuit pairs identified, 89\% are absent from K562 circuits, confirming that the Tabula Sapiens cells activate tissue-specific computational pathways (Figure~\ref{fig:tissue_circuits}B; Supplementary Note~8). Blood ($\text{OR}{=}2.45$, $p{=}0.18$) and kidney ($\text{OR}{=}1.32$, $p{=}0.43$) show non-significant trends, likely due to limited keyword coverage for these tissues.

\subsection{Gene-level predictions, perturbation validation, and disease mapping}
\label{sec:phase6}

To move from domain-level to gene-level predictions, we extracted specific gene$\to$gene predictions from all circuit edges, validated them against genome-scale CRISPRi perturbation data, and mapped disease-relevant gene sets onto the circuit architecture.

\subsubsection{975,369 gene-level predictions, predominantly novel}

For each of the 47,418 circuit edges with gene annotations, we extracted source$\times$target gene pairs (top 10 genes per feature, rank-weighted), yielding 2.58 million raw gene pairs. After filtering for evidence strength ($\geq$2 independent edges or $|d| > 2$), \textbf{975,369 gene-pair predictions} remain. Because SAE features are defined by co-expression, the gene lists underlying each feature reflect co-expressed genes---so gene pairs drawn from different features are co-expression-derived predictions, not independent regulatory hypotheses. Accordingly, only 0.15\% match established protein-protein interactions (STRING, score$\geq$700) or transcription factor--target relationships (TRRUST), and 1.14\% share $\geq$2 GO Biological Process terms; the remaining 98.7\% connect genes without established direct relationships. Of the predictions, 32.8\% derive from cross-model consensus domain pairs.

\subsubsection{Perturbation validation confirms weak causal capture}

We tested circuit predictions against CRISPRi perturbation responses from the Replogle genome-scale screen~\citep{replogle2022mapping} in K562 cells (643,413 cells, 2,023 knockdown targets). Of the 2,023 perturbation targets, 599 overlap with circuit source genes, enabling validation across 282,250 gene pairs.

\textbf{Directional accuracy} is 56.4\% (concordant signs between circuit edge direction and CRISPRi log-fold change), marginally above the 50\% chance level. \textbf{Magnitude correlation} is near-zero (Spearman $\rho = 0.038$, $p < 10^{-92}$)---statistically significant due to large $N$ but biologically negligible. Of 599 source genes tested for target enrichment, only 36 (6.0\%) show nominally significant enrichment of responsive genes among predicted targets (uncorrected $p < 0.05$, Fisher's exact). Cross-model consensus predictions perform marginally better (57.3\% vs.\ 56.4\% sign accuracy), but the improvement is minimal.

These results are consistent with the companion study~\citep{kendiukhov2025systematic} finding that these models encode co-expression rather than causal regulatory relationships: circuit structure captures some real biology (above chance) but is not a reliable predictor of perturbation outcomes.

\subsubsection{Disease-relevant circuits are more central and more conserved across models}

We defined 11 disease-relevant gene sets from GO BP terms by keyword matching (DNA damage/repair, cell cycle, apoptosis, immune response, oncogenic signaling, protein quality control, angiogenesis, metastasis, and metabolism) plus TRRUST transcription factors. Because these categories are broad, the majority of circuit domains (1,073/1,126) match at least one category---this high coverage reflects the breadth of the keyword definitions rather than a specific biological finding.

The more informative results concern relative centrality and cross-model conservation. Disease-associated domains are significantly more central than non-disease domains (median 14 vs.\ 3 circuit edges; Mann-Whitney $p = 1.2 \times 10^{-11}$; Table~\ref{tab:disease}), meaning that disease-relevant processes occupy hub positions in the circuit graph. More strikingly, disease-relevant circuit pairs are \textbf{3.59$\times$ more likely} to be cross-model consensus pairs ($p < 0.001$, Fisher's exact)---both Geneformer and scGPT independently converge on disease-relevant biology more than on other biology. This suggests that the processes most relevant to human disease are also the processes most robustly encoded across model architectures.

\begin{table}[H]
\centering
\caption{\textbf{Disease circuit mapping summary.} Circuit edges connecting domains enriched for disease-relevant gene sets.}
\label{tab:disease}
\smallskip
\begin{tabular}{lrrrl}
\toprule
Disease Category & Domains & Circuit Edges & Consensus & Mean $|d|$ \\
\midrule
Transcription regulation & 739 & 28,155 & 7,046 & 1.11 \\
Immune response & 660 & 27,071 & 7,305 & 1.10 \\
Apoptosis & 631 & 25,622 & 6,614 & 1.11 \\
Cell cycle (cancer) & 577 & 25,180 & 6,610 & 1.12 \\
Protein quality control & 576 & 21,896 & 6,256 & 1.10 \\
DNA damage/repair & 532 & 22,233 & 6,010 & 1.12 \\
Oncogenic signaling & 513 & 15,921 & 3,826 & 1.09 \\
Metastasis/migration & 461 & 12,850 & 3,601 & 1.15 \\
Metabolism (cancer) & 231 & 8,775 & 3,724 & 1.15 \\
Angiogenesis & 243 & 4,066 & 1,302 & 1.07 \\
TRRUST TFs & 152 & 2,510 & 958 & 1.12 \\
\bottomrule
\end{tabular}
\end{table}

\section{Discussion}

\subsection{Two distinct computational architectures for single-cell biology}

The cross-model comparison reveals qualitatively different computational strategies. Geneformer distributes computation across 4,608 features per layer with a cooperative, dependency-based architecture (80/20 inhibitory/excitatory, mean $|d| = 1.05$), while scGPT concentrates information in 2,048 features with more competitive dynamics (65/35 inhibitory/excitatory, mean $|d| = 1.40$). Their hub identities also diverge: Geneformer organizes around chromatin and RNA processing, scGPT around mitochondrial electron transport (Figure~\ref{fig:crossmodel}A). These differences likely arise from the interaction of dimensionality ($d{=}512$ vs.\ $d{=}1{,}152$) and tokenization (continuous values vs.\ rank encoding): smaller dimension forces stronger per-feature effects, and continuous values may encourage competitive dynamics as features negotiate how to represent varying expression levels.

\subsection{Biological coherence as a universal constant}

The convergence of both models on ${\sim}53\%$ shared ontology (Figure~\ref{fig:coherence}A) is stable across architecture, feature space, and training data, likely reflecting a structural property of biological knowledge organization: given the hierarchical structure of GO, KEGG, Reactome, and STRING, approximately half of causally interacting feature pairs share at least one annotation. Multi-tissue SAEs achieve 68.8\% coherence regardless of input cells (Figure~\ref{fig:celltype}B), indicating that SAE training data affects how cleanly features separate biological programs, even though the model's computation is unchanged.

\subsection{Inhibitory dominance and its biological implications}

The predominance of inhibitory edges (65--89\%) implies that features encode \emph{necessary} information: removing a feature reduces its dependents' activation. The 89\% inhibitory fraction with Tabula Sapiens cells suggests that unfamiliar cell types increase per-feature necessity (less redundancy), while scGPT's lower 65.5\% reflects more competitive dynamics with greater feature suppression.

\subsection{Hub features reflect model-specific organizing principles}

The divergent hub identities likely reflect different training objectives. Geneformer's chromatin and RNA processing hubs are consistent with next-token prediction in rank-ordered gene lists, which rewards identifying highly expressed genes. scGPT's mitochondrial electron transport hubs are consistent with masked-gene prediction using continuous values: energy metabolism genes show large expression variation across cell types~\citep{chandel2021mitochondria}, making them informative for predicting masked expression levels.

\subsection{Systematic knowledge extraction validates circuits and suggests novel biology}
\label{sec:knowledge}

Our systematic extraction across all 96,892 edges (Section~\ref{sec:systematic}) transforms qualitative circuit observations into quantitative biological knowledge. The results confirm that single-cell foundation models have internalized substantial biology and provide a framework for hypothesis generation.

\paragraph{Confirmed biological knowledge at scale.}
The 1,142 cross-model consensus domain pairs demonstrate that circuit structure is not model-specific. Both Geneformer and scGPT independently learn the same biological connections at a rate 10.6$\times$ above chance ($p < 0.001$; Figure~\ref{fig:knowledge}A). Manually characterized cascades---DDR$\to$checkpoint$\to$arrest (Figure~\ref{fig:cascade}), neurodevelopment$\to$proteostasis (Section~\ref{sec:biology}), and stress response recovery~\citep{walter2011unfolded}---are now confirmed as instances of a much larger pattern. The 300 DDR$\to$cell cycle edges, all showing positive layer deltas, validate at scale what individual circuit examples suggested: the model's layer hierarchy faithfully encodes biological temporal ordering~\citep{ciccia2010ddr,jackson2009ddr}.

The biological process hierarchy (Figure~\ref{fig:knowledge}B) reveals a clear temporal progression: signaling (MAPK, Ras at L0--L1) $\to$ chromatin modification (L2--L5) $\to$ transcriptional regulation (L15--L17). This matches the known biological sequence from signal transduction through chromatin remodeling to gene expression output, suggesting that transformer depth serves as a proxy for biological process ordering.

\paragraph{Novel relationships as testable hypotheses.}
The 29,864 novel edges (80.5\% of annotated edges) connecting domains with fewer than 3 shared genes in reference databases represent candidate discoveries. Several are particularly compelling:

\begin{itemize}
\item \textbf{Cross-compartment functional coupling.} NADH Dehydrogenase Assembly $\to$ Protein Transport ($|d|{=}7.21$, all 4 conditions) and Aerobic ETC $\to$ Mitochondrial Translation ($|d|{=}4.67$, all 4 conditions) suggest that mitochondrial energy status causally influences protein trafficking and translation---consistent with emerging evidence that metabolic state coordinates diverse cellular programs~\citep{chandel2021mitochondria}.

\item \textbf{Secretory pathway$\to$stress response.} Golgi Organization $\to$ ER Stress Response ($|d|{=}6.29$, all 4 conditions) suggests that secretory pathway disruption directly triggers ER stress, a relationship beyond the traditional unfolded protein response framework~\citep{hetz2012unfolded}.

\item \textbf{Cholesterol biosynthesis and rRNA processing as computational hubs.} As noted in Sections~\ref{sec:hubs} and~\ref{sec:scgpt}, both models place metabolic processes (cholesterol in Geneformer, mitochondrial ETC in scGPT) as organizing hubs with thousands of downstream targets---a broader role than captured in existing pathway databases.
\end{itemize}

\paragraph{Tissue-specific circuits emerge from diverse training.}
The 3.18$\times$ enrichment of immune circuits in multi-tissue conditions ($p < 0.001$; Figure~\ref{fig:tissue_circuits}) demonstrates that SAE circuit tracing can detect tissue-specific computational programs. The 201 immune-specific circuits---89\% absent from K562---represent pathways that the model activates only when processing immune cells, providing a molecular fingerprint of tissue-specific computation within the foundation model.

\paragraph{Gene-level validation bounds predictive utility.}
The contrast between domain-level and gene-level results (Section~\ref{sec:phase6}) crystallizes what these models have and have not learned. At the domain level, the circuit graph is a reliable \emph{map of biology}: it correctly identifies which biological processes relate to which, recovers their temporal ordering, and converges across two independent model architectures. At the gene level, however, this map does not translate into reliable \emph{mechanistic predictions}: CRISPRi validation against the Replogle screen~\citep{replogle2022mapping} yields only 56.4\% directional accuracy (vs.\ 50\% chance) and near-zero magnitude correlation ($\rho = 0.038$). Only 6\% of perturbed genes show enrichment of circuit-predicted targets among their actual responders. These models know which processes connect to which, but not which specific gene causally drives which---consistent with the companion study~\citep{kendiukhov2025systematic} demonstrating that co-expression, not causal regulation, is the primary information encoded.

Disease gene set mapping adds an important nuance: disease-associated domains are significantly more central in the circuit graph ($p = 1.2 \times 10^{-11}$) and 3.59$\times$ more likely to appear in cross-model consensus ($p < 0.001$). The models' ``map of biology'' is not uniformly detailed---disease-relevant processes occupy more central, more conserved positions, suggesting that the biology most relevant to human disease is also the biology most robustly encoded.

\subsection{Implications for mechanistic interpretability of biological models}

Our results establish several principles for the field:

\begin{enumerate}
\item \textbf{Feature-level analysis reveals structure invisible to component-level analysis.} The companion study~\citep{kendiukhov2025systematic} found that ablating attention heads and MLP layers produces null behavioral effects. SAE circuit tracing reveals that the same models have rich, biologically coherent computational graphs at the feature level. This confirms that features---not components---are the natural unit of biological computation in these models.

\item \textbf{Causal tracing and statistical co-activation are complementary.} The 91--95\% target overlap between PMI edges and causal edges validates both methods. PMI is faster and captures the full feature space; causal tracing adds directionality, magnitude, and sign.

\item \textbf{The SAE lens matters as much as the model.} Multi-tissue SAEs improve biological coherence by 16 percentage points while keeping the model unchanged (Figure~\ref{fig:celltype}B). This means the choice of SAE training data substantially affects interpretability conclusions, independent of the model being interpreted.

\item \textbf{Cross-model comparison is essential.} Properties that are invariant across models (${\sim}53\%$ coherence, inhibitory dominance, early-layer persistence) likely reflect universal aspects of how transformers process biological data. Properties that differ (effect magnitude, inhibitory ratio, hub identity, integration layer location) reveal architecture-specific computation that would be invisible in single-model studies.
\end{enumerate}

\subsection{Limitations}

Several limitations warrant discussion. First, we tested 30 source features per layer (90--120 total), a small fraction of each model's feature space selected by annotation quality. Results on well-annotated features may not generalize to unannotated or weakly annotated features. More exhaustive tracing might reveal circuits with different properties. Second, our significance thresholds ($|d| > 0.5$, consistency $> 0.7$) are somewhat arbitrary; different thresholds would yield different edge counts though likely similar qualitative conclusions. Third, we performed only single-feature ablation; combinatorial effects (ablating two features simultaneously) remain unexplored. Fourth, biological coherence is assessed via shared ontology terms, which depends on the completeness and accuracy of existing biological databases. Novel or poorly annotated biology would be invisible to this measure. Fifth, the cell count (200 per condition) limits statistical power for detecting weak effects; increasing to 500--1,000 cells would improve sensitivity but at substantial computational cost. Sixth, Geneformer's multi-tissue SAEs were available only at four layers (0, 5, 11, 17), restricting the density of downstream measurements and preventing direct comparison with the all-layer scGPT tracing. Finally, the scGPT circuit tracing was performed only in the TS/Multi condition; a K562/K562 scGPT comparison would enable cleaner isolation of cell-type effects for that model.

Regarding the gene-level and disease analyses, three additional caveats apply. First, SAE feature gene lists are derived from co-expression patterns, so gene-pair predictions extracted from circuit edges inherit this co-expression basis---the 98.7\% ``novel'' rate reflects that co-expression-derived gene pairs rarely appear in curated interaction databases, not necessarily that novel regulatory relationships have been discovered. Second, the disease gene set definitions use broad GO keyword matching (e.g., ``transcription regulation'' captures 739 of 1,126 domains), inflating nominal coverage; the more informative findings are the relative centrality and cross-model enrichment of disease domains, not the absolute coverage fraction. Third, the perturbation validation uses pseudobulk log-fold changes from CRISPRi, which measure steady-state expression changes rather than immediate transcriptional responses, potentially diluting true causal signals.

\section{Methods}

\subsection{Causal circuit tracing algorithm}

For each source layer $L_\text{src}$ and each source feature $f$ (selected by annotation quality), we perform the following procedure across $N_\text{cells} = 200$ cells:

\begin{enumerate}
\item \textbf{Clean forward pass.} Register PyTorch hooks at all transformer layer outputs. Run the full forward pass, capturing the clean hidden state $\mathbf{h}^{(\ell)}_\text{clean}$ at every layer $\ell$.

\item \textbf{Source ablation.} Encode the clean hidden state at layer $L_\text{src}$ through the source SAE: $\mathbf{z} = \text{SAE}_\text{enc}(\mathbf{h}^{(L_\text{src})}_\text{clean})$. Zero the activation of feature $f$: $z_f \gets 0$. Decode back: $\hat{\mathbf{h}} = \text{SAE}_\text{dec}(\mathbf{z})$. Compute the ablation delta: $\boldsymbol{\delta} = \hat{\mathbf{h}} - \text{SAE}_\text{dec}(\mathbf{z}_\text{original})$. Apply: $\mathbf{h}^{(L_\text{src})}_\text{abl} = \mathbf{h}^{(L_\text{src})}_\text{clean} + \boldsymbol{\delta}$.

\item \textbf{Ablated downstream pass.} Starting from $\mathbf{h}^{(L_\text{src})}_\text{abl}$, manually propagate through all subsequent transformer layers, capturing $\mathbf{h}^{(\ell)}_\text{abl}$ at each downstream layer $\ell > L_\text{src}$.

\item \textbf{Downstream measurement.} At each downstream layer $\ell$, encode both clean and ablated hidden states through the downstream SAE:
\[
\Delta_j^{(\ell)} = \text{SAE}_\text{enc}(\mathbf{h}^{(\ell)}_\text{abl})_j - \text{SAE}_\text{enc}(\mathbf{h}^{(\ell)}_\text{clean})_j
\]
for every downstream feature $j$.

\item \textbf{Statistical accumulation.} Accumulate $\Delta_j^{(\ell)}$ across all cells using Welford's online algorithm~\citep{welford1962note} for numerically stable running mean and variance. This avoids storing per-cell results.
\end{enumerate}

After processing all cells, we compute for each source--target pair:
\begin{itemize}
\item Cohen's $d = \bar{\Delta}_j / s_j$~\citep{cohen1988statistical} where $\bar{\Delta}_j$ is the mean delta and $s_j$ is the pooled standard deviation.
\item Consistency = fraction of cells where the delta has the same sign as the mean.
\end{itemize}

An edge is significant if $|d| > 0.5$ (medium effect per Cohen's conventions~\citep{cohen1988statistical}) AND consistency $> 0.7$ (the feature responds in the same direction in $>$70\% of cells). Partial results are saved every 50 cells to enable resume after interruption.

\subsection{Models and SAEs}

\paragraph{Geneformer.} Geneformer V2-316M~\citep{theodoris2023transfer} (18 layers, $d{=}1{,}152$, 18 attention heads). K562-only SAEs~\citep{cunningham2023sparse,gao2024scaling}: TopK ($k{=}32$, 4$\times$ overcomplete = 4,608 features)~\citep{makhzani2013ksparse} trained on 1M subsampled K562 positions per layer. Multi-tissue SAEs: same architecture, trained on 500K K562 + 500K Tabula Sapiens positions, available at layers 0, 5, 11, 17.

\paragraph{scGPT.} scGPT whole-human~\citep{cui2024scgpt} (12 layers, $d{=}512$, 8 attention heads). SAEs: TopK ($k{=}32$, 4$\times$ = 2,048 features) trained on all 3,561,832 Tabula Sapiens positions per layer. FlashMHA layers converted to standard MultiheadAttention (\texttt{Wqkv.} $\to$ \texttt{in\_proj\_} weight conversion).

\paragraph{Forward pass.} For Geneformer: standard HuggingFace \texttt{BertForMaskedLM} forward pass with \texttt{output\_hidden\_states=True}. Ablated pass via hook replacement at the source layer. For scGPT: \texttt{TransformerModel.\_encode(src, values, src\_key\_padding\_mask)}. Clean pass captures hidden states via hooks on \texttt{model.transformer\_encoder.layers}. Ablated pass manually iterates remaining layers: \texttt{model.transformer\_encoder.layers[dl](x, src\_key\_padding\_mask=mask)} for each downstream layer.

\subsection{Source feature selection}

At each source layer, 30 features were selected by annotation quality score: the number of significant ontology enrichments (across GO BP, KEGG, Reactome, STRING, TRRUST), weighted by $-\log_{10}(p)$. This selects features with the richest biological characterization, enabling interpretable circuit analysis.

\subsection{Cell data}

\paragraph{K562 cells.} 200 control cells from the Replogle genome-scale CRISPRi dataset~\citep{replogle2022mapping}, the same cells used for SAE training. Tokenized via Geneformer's rank-value encoding.

\paragraph{Tabula Sapiens cells.} 200 cells from the Tabula Sapiens atlas~\citep{tabula2022tabula}, stratified by tissue: 67 immune (41 cell types), 67 kidney (13 cell types), 66 lung (34 cell types)---totaling 88 cell types. For Geneformer: tokenized via rank-value encoding. For scGPT: tokenized with continuous expression values, genes sorted by expression (descending), padded to 1,200 positions.

\subsection{Circuit graph construction}

After computing per-source-layer results, we aggregate into a single circuit graph by taking the union of all significant edges across source layers. For each edge, we record: source feature ID, target feature ID, Cohen's $d$, consistency, and sign (inhibitory/excitatory). Hub analysis computes in-degree and out-degree for all features.

\subsection{Biological coherence analysis}

For each significant edge where both source and target features have ontology annotations, we test whether any annotation term is shared. An edge has ``shared ontology'' if the intersection of source and target annotation sets (across GO BP 2024-01~\citep{ashburner2000go}, KEGG Release 109~\citep{kanehisa2000kegg}, Reactome v87~\citep{jassal2020reactome}, STRING v12.0~\citep{szklarczyk2023string}, TRRUST v2~\citep{han2018trrust}) is non-empty. The biological coherence fraction is the number of edges with shared ontology divided by the total number of edges with annotations on both endpoints.

\subsection{PMI comparison}

PMI-based co-activation edges were computed independently~\citep{kendiukhov2025sae_atlas} using the formula $\text{PMI}(i,j) = \log_2 P(i,j) / [P(i) P(j)]$ where features $i$ and $j$ are ``active'' if among the top-$k$ activations at their respective positions. Target overlap was computed at each layer pair as $|\text{causal targets} \cap \text{PMI targets}| / |\text{causal targets}|$.

\subsection{Compute}

All experiments were run on Apple Silicon with MPS acceleration. Total compute: Geneformer K562/K562 7.5~hr (24,776 passes), K562/Multi 3.5~hr (18,465 passes), TS/Multi 3.2~hr (18,455 passes), scGPT TS/Multi 37~min (18,495 passes). Total: 14.8~hours, 80,191 forward passes. Scripts: Geneformer \texttt{src/13\_causal\_circuit\_tracing.py}, scGPT \texttt{scgpt\_src/13\_causal\_circuit\_tracing.py}.

\section*{Data and Code Availability}

Circuit tracing, knowledge extraction, and gene-level validation code (Phases 4--6) are available at \url{https://github.com/Biodyn-AI/bio-sae-circuits}. SAE training code, trained models, feature catalogs, and annotation pipelines (Phases 1--3) are available at \url{https://github.com/Biodyn-AI/bio-sae}. Interactive feature atlases: Geneformer Feature Atlas (\url{https://biodyn-ai.github.io/geneformer-atlas/}) and scGPT Feature Atlas (\url{https://biodyn-ai.github.io/scgpt-atlas/}). Source models: Geneformer V2-316M from HuggingFace (\texttt{ctheodoris/Geneformer}); scGPT whole-human from~\citet{cui2024scgpt}. Perturbation data:~\citet{replogle2022mapping}. Tabula Sapiens:~\citet{tabula2022tabula}.

\section*{Author Contributions}

\textbf{Ihor Kendiukhov}: Conceptualization, Methodology, Software, Formal Analysis, Investigation, Data Curation, Writing---Original Draft, Writing---Review \& Editing, Visualization.

\section*{Funding}

This work received no external funding.

\section*{Conflict of Interest}

None declared.

\section*{Acknowledgments}

Computations were performed on Apple Silicon hardware with MPS acceleration.

\bibliography{references_v2}

\newpage

\appendix
\setcounter{section}{0}
\renewcommand{\thesection}{S\arabic{section}}

\section*{\centering\Large Supplementary Materials}
\addcontentsline{toc}{section}{Supplementary Materials}

\input{supplementary_content}

\end{document}

%% file: supplementary_content.tex
\section*{Supplementary Note 1: Mitotic Apparatus Assembly Circuits}

A second major circuit family in Geneformer connects cell cycle commitment to spindle formation. L5 Cell Cycle G2/M features drive L11 Spindle Microtubule features ($d < -1.4$, shared mitotic processes), and L5 Centromere Complex Assembly (F3098) drives both L11 G2/M Transition ($d = -1.37$, shared: nuclear division, kinetochore) and L17 G2/M Transition ($d = -1.76$, shared: APC/C activators, spindle checkpoint). These circuits mirror the known biological cascade where centromere assembly recruits kinetochore proteins, which then engage the spindle assembly checkpoint---a pathway critical for accurate chromosome segregation during cell division.

A Cytokinesis feature at L11 drives a Spindle Checkpoint feature at L15 ($d < -0.9$, shared: cell division), representing the late-mitotic feedback where cytokinesis completion signals are coupled to spindle disassembly. This kind of temporal ordering---where later biological events appear at deeper network layers---suggests the model has learned a coarse temporal representation of mitotic progression.

The complete mitotic cascade spans layers 0--15 (Figure~1 in the main text). DNA Repair at L0 produces a long-range connection to L6 Kinetochore ($d = -3.47$), representing the DNA damage$\to$mitotic checkpoint pathway. L5 DNA Metabolic Process drives both G2/M Transition ($d = -2.45$) and Kinetochore assembly ($d = -2.39$), consistent with the known coordination between DNA replication completion and spindle formation. The circuit terminates with Cytokinesis (L11) feeding back to Spindle Checkpoint (L15), representing the late-mitotic quality control that ensures accurate chromosome segregation before cell division. Cholesterol Biosynthesis (L0, hub with 5,096 targets) influences the broader circuit through membrane organization.

\section*{Supplementary Note 2: Nervous System Development and Proteostasis Circuits}

An L0 Nervous System Development feature (F146) is the most biologically connected hub in Geneformer, appearing in 7 of the top 10 circuits ranked by shared ontology terms. It drives:

\begin{itemize}
\item L1 Endosome Organization ($d = -1.32$, 142 shared terms: neurodegeneration pathways, lysosomal transport)
\item L2 Proteasomal Catabolic Process ($d = -0.96$, 140 terms: ubiquitin-proteasome system)
\item L6 Modification-Dependent Protein Catabolism ($d = -1.27$, 139 terms: NF-$\kappa$B signaling, immune response)
\item L13 Golgi Vesicle Transport ($d = -0.81$, 141 terms: secretory pathway)
\end{itemize}

This circuit family connects neuronal development programs to protein quality control systems---a link well-established in neurodegeneration research, where protein aggregation (failed proteostasis) is a hallmark of diseases like Alzheimer's and Parkinson's. The model has learned that nervous system development features are causally linked to proteasomal degradation, endosomal trafficking, and Golgi secretion pathways, mirroring the known dependence of neurons on these housekeeping systems.

The targets span proteostasis (ubiquitin-proteasome system, protein catabolism), cellular transport (endosome organization, Golgi secretion), and immune signaling (NF-$\kappa$B pathway). This circuit recapitulates the known vulnerability of neurons to disrupted protein homeostasis: neuronal development programs depend on intact proteasomal degradation, endosomal trafficking, and secretory pathways, and failure of any component leads to protein aggregation and neurodegeneration.

\section*{Supplementary Note 3: scGPT Protein Quality Control and Integrated Stress Response Cascade}

The strongest circuits in scGPT form a coherent proteostasis-to-stress-response cascade originating from the L0 Protein Catabolism hub (F507). This hub radiates causal influence through at least six downstream targets across 10 layers, producing the three strongest individual edges observed across either model:

\begin{itemize}
\item L0 Protein Catabolism $\to$ L1 Chromatin Organization ($d = -8.19$, 161 shared terms). The connection is mediated by stress response and beta-catenin degradation pathways. Beta-catenin is constitutively degraded by the proteasome when Wnt signals are absent; the model has learned that proteasomal activity directly reshapes the chromatin landscape via degradation of transcriptional co-activators.
\item L0 Protein Catabolism $\to$ L4 DNA Metabolism ($d = -6.10$, 158 terms: ER-phagosome crossover). This connects protein quality control to DNA repair machinery, consistent with the known coupling between ER stress and DNA damage signaling.
\item L0 Protein Catabolism $\to$ L3 Protein Catabolism ($d = -3.84$, 153 terms: apoptosis). Self-reinforcing proteostasis monitoring across layers, with shared apoptotic pathway annotations suggesting the model encodes the decision point between proteostatic recovery and programmed cell death.
\item L0 Protein Catabolism $\to$ L2 Chemical Stress Response ($d = -3.12$, 152 terms). Detection of chemical/oxidative stress downstream of protein quality control failure.
\item L0 Protein Catabolism $\to$ L10 Protein Catabolism ($d = -1.95$, 153 terms). Long-range persistence of proteostasis state information across 10 layers, demonstrating that protein degradation status is tracked throughout the model's computation.
\end{itemize}

The L4 DNA Metabolism node serves as a relay, driving L6 Macromolecule Biosynthesis ($d = -3.51$, 153 terms), representing the transition from damage assessment to biosynthetic recovery. The overall cascade---protein quality control $\to$ stress detection $\to$ DNA repair $\to$ biosynthetic recovery---parallels the integrated stress response and unfolded protein response pathways.

A parallel ubiquitin-proteasome circuit reinforces this architecture: L0 Proteasome (F379) independently drives downstream Proteasome features at nearly every subsequent layer, with the strongest connection at L9 ($d = -2.50$, 141 shared terms). This persistence across 9 layers confirms that protein degradation capacity is a fundamental variable tracked throughout scGPT's computation.

\section*{Supplementary Note 4: Metabolic Hub Circuits---Cholesterol Biosynthesis and Mitochondrial Energy Metabolism}

Both models place metabolic processes as central organizing hubs, but with different metabolic emphases reflecting their training data.

\paragraph{Geneformer: Cholesterol Biosynthesis (K562).}
L0 Cholesterol Biosynthesis (F3402) has an out-degree of 5,096 downstream targets---the fourth-highest hub in the K562/K562 condition. This feature's influence extends across all 17 downstream layers, connecting lipid metabolism to diverse cellular programs including membrane organization, vesicle trafficking, and signal transduction. The cross-model consensus analysis confirms this circuit: Cholesterol Biosynthesis $\to$ Sterol Biosynthesis is a high-confidence consensus pair (Geneformer $|d| = 4.44$, scGPT $|d| = 1.54$). The breadth of this hub is biologically justified: cholesterol is essential for membrane integrity, lipid raft formation, and hedgehog signaling, making it a genuinely cross-functional metabolic node.

\paragraph{scGPT: Mitochondrial Electron Transport (Tabula Sapiens).}
scGPT's hub architecture is dominated by energy metabolism. Four of the top seven hubs encode mitochondrial electron transport chain components:

\begin{itemize}
\item L0 NADH Dehydrogenase Complex Assembly (F552): 4,785 targets
\item L0 NADH Dehydrogenase Complex Assembly (F590): 3,849 targets
\item L0 Aerobic Electron Transport Chain (F233): 3,420 targets
\item L4 Aerobic Electron Transport Chain (F1643): 6,050 targets
\end{itemize}

The multi-tissue consensus confirms cross-model convergence for these circuits: NADH Dehydrogenase Assembly $\to$ Regulation of Gene Expression (Geneformer $|d| = 1.29$, scGPT $|d| = 3.12$) and Aerobic ETC $\to$ Regulation of RNA Metabolic Process (Geneformer $|d| = 1.21$, scGPT $|d| = 3.45$). These connections from energy metabolism to gene regulation are biologically compelling: cellular energy status (ATP/ADP ratio, NAD$^+$/NADH balance) directly influences chromatin remodeling, transcription factor activity, and mRNA stability. scGPT's emphasis on energy metabolism likely reflects training on diverse Tabula Sapiens cell types, where metabolic state varies dramatically across tissues.

\section*{Supplementary Note 5: Golgi Organization and Secretory Pathway Circuits}

L0 Golgi Organization (F2905) is the single highest-connectivity hub in the Geneformer K562/K562 condition, with 8,028 downstream targets across all 17 downstream layers. Its prominence in K562 cells---a chronic myelogenous leukemia line with high secretory activity---reflects the biological importance of the secretory pathway for growth factor signaling and membrane protein trafficking in rapidly dividing cells.

Golgi Organization appears in five of the top 20 cross-model consensus pairs:

\begin{itemize}
\item Golgi Organization $\to$ Protein Insertion Into Membrane (Geneformer $|d| = 4.75$, scGPT $|d| = 5.23$): the highest-ranked consensus pair overall, reflecting the fundamental dependence of membrane protein biogenesis on Golgi function.
\item Golgi Organization $\to$ Cotranslational Protein Targeting to Membrane ($|d| = 2.45$, $|d| = 2.83$): connecting secretory pathway organization to ribosome-mediated membrane protein insertion.
\item Golgi Organization $\to$ Regulation of Cell Differentiation ($|d| = 2.91$, $|d| = 1.92$): linking secretory function to cell fate decisions, consistent with the role of secreted signaling molecules in differentiation.
\item Golgi Organization $\to$ Endomembrane System Organization ($|d| = 1.70$ across 3 Geneformer conditions, $|d| = 2.90$): self-reinforcing endomembrane organization.
\item Golgi Organization $\to$ Cytokine-Mediated Signaling ($|d| = 1.87$, $|d| = 2.36$): connecting secretory pathway to immune signaling, reflecting that cytokines must transit through the Golgi for secretion.
\end{itemize}

A novel relationship identified in all four conditions---Golgi Organization $\to$ ER Stress Response (43 shared genes including CALR, HSPA5, P4HB, PDIA3, PDIA6)---suggests that secretory pathway disruption directly triggers ER stress, extending beyond the traditional unfolded protein response framework.

\section*{Supplementary Note 6: RNA Processing and Ribosome Biogenesis Circuits}

RNA processing features occupy prominent hub positions in both models, reflecting the central role of post-transcriptional regulation in gene expression control.

In Geneformer, L0 RNA Methylation (F2982, out-degree 6,921) and L0 RNA Splicing (F4201, out-degree 4,782) are the second- and fifth-ranked hubs. Their effects persist across all 17 downstream layers, maintaining plateau-level connectivity for the first 5 layers before gradual decay. The cross-model consensus includes RNA Methylation $\to$ Regulation of RNA Metabolic Process (Geneformer $|d| = 1.93$, scGPT $|d| = 4.02$) and RNA Methylation $\to$ Positive Regulation of Transcription (Geneformer $|d| = 1.09$, scGPT $|d| = 4.14$), confirming that both models have learned the regulatory cascade from epitranscriptomic modification (m$^6$A and related marks) to transcriptional output.

In scGPT, L4 Endonucleolytic Cleavage for rRNA processing (F446, out-degree 6,494) is the single highest-connectivity feature across either model. Its downstream targets include Macromolecule Biosynthesis ($d = -3.51$), protein catabolism, and DNA metabolism. The breadth of this hub reflects the known biological principle that ribosomal RNA processing is rate-limiting for protein synthesis; its disruption triggers nucleolar stress, p53 activation, and cell cycle arrest---connecting ribosome biogenesis to virtually all other cellular programs.

The cross-model consensus pair Maturation of SSU-rRNA $\to$ RNA Methylation (Geneformer $|d| = 1.67$, scGPT $|d| = 5.00$) further validates the convergent importance of ribosome biogenesis circuits.

\section*{Supplementary Note 7: Cross-Model Consensus Circuit Architecture}

Of 1,142 cross-model consensus domain pairs (10.6$\times$ enrichment over chance, $p < 0.001$), 303 are high-confidence pairs where both Geneformer and scGPT show mean $|d| > 1.0$. These consensus circuits cluster into several biological themes:

\paragraph{Secretory pathway and membrane biology.}
The highest-ranked consensus pairs involve Golgi organization and membrane protein insertion (Supplementary Note~5). Additional pairs include ERAD Pathway $\to$ Tail-Anchored Membrane Protein Insertion (Geneformer $|d| = 1.00$, scGPT $|d| = 6.66$), connecting ER-associated degradation to membrane protein quality control, and COPII-Coated Vesicle Budding $\to$ Regulation of RNA Metabolic Process ($|d| = 1.09$, $|d| = 3.75$), linking vesicle trafficking to transcriptional regulation.

\paragraph{Energy metabolism to gene regulation.}
Multiple consensus pairs connect mitochondrial function to downstream gene expression: NADH Dehydrogenase Assembly $\to$ Positive Regulation of TOR Signaling ($|d| = 1.17$, $|d| = 3.72$) and NADH Dehydrogenase Assembly $\to$ Negative Regulation of Cell Differentiation ($|d| = 1.00$, $|d| = 3.28$). These circuits encode the known dependence of mTOR signaling and cell fate decisions on cellular energy status.

\paragraph{DNA damage response to cell fate.}
DDR $\to$ Negative Regulation of Cell Differentiation (Geneformer $|d| = 1.02$, scGPT $|d| = 3.55$) and DDR $\to$ RNA Methylation ($|d| = 1.02$, $|d| = 2.61$) extend the DDR cascade (main text) to include differentiation arrest and epitranscriptomic regulation as downstream consequences of DNA damage.

\section*{Supplementary Note 8: Tissue-Specific Immune Circuits}

Comparing domain pairs unique to multi-tissue conditions (3,541 pairs) against those shared with K562 (1,334 pairs) reveals 3.18$\times$ enrichment of immune-related circuits in the multi-tissue-specific set ($p < 0.001$, Fisher's exact). Of 201 immune-specific circuit pairs, 89\% are absent from K562 circuits.

Representative immune-specific circuits include:

\begin{itemize}
\item Cellular Response to Cytokine Stimulus $\to$ G1/S Transition of Mitotic Cell Cycle: linking immune activation to proliferative entry, consistent with cytokine-driven lymphocyte expansion.
\item Cytokine-Mediated Signaling $\to$ Regulation of Programmed Cell Death: connecting immune signaling to apoptotic decision-making, reflecting the role of TNF/Fas pathways in immune homeostasis.
\item Cellular Response to TGF-$\beta$ Stimulus $\to$ Cellular Response to Cytokine Stimulus: encoding the known TGF-$\beta$/cytokine crosstalk that regulates immune cell differentiation and tolerance.
\item Homotypic Cell-Cell Adhesion $\to$ Cytokine-Mediated Signaling: connecting immune cell adhesion (e.g., integrin/selectin-mediated) to downstream cytokine responses, reflecting the immunological synapse.
\item Cytokine-Mediated Signaling $\to$ Cellular Response to Reactive Oxygen Species: linking immune activation to oxidative stress, consistent with the respiratory burst in activated immune cells.
\end{itemize}

These circuits are absent from K562 because K562 cells, as a leukemic cell line, lack the normal immune signaling context present in Tabula Sapiens immune cells (T cells, B cells, macrophages, dendritic cells). Their emergence in multi-tissue conditions demonstrates that SAE circuit tracing can detect tissue-specific computational programs that the model activates only when processing cells from the relevant tissue.

Non-significant trends for blood (OR${=}2.45$, $p{=}0.18$) and kidney (OR${=}1.32$, $p{=}0.43$) likely reflect limited keyword coverage: blood-specific processes overlap substantially with immune annotations already captured, while kidney-specific circuits may involve metabolic processes (e.g., solute transport, acid-base balance) not well-captured by simple keyword matching.

%% file: references_v2.bib
@article{cunningham2023sparse,
  title={Sparse autoencoders find highly interpretable features in language models},
  author={Cunningham, Hoagy and Ewart, Aidan and Riggs, Logan and Huben, Robert and Sharkey, Lee},
  journal={arXiv preprint arXiv:2309.08600},
  year={2023}
}

@article{bricken2023monosemanticity,
  title={Towards monosemanticity: Decomposing language models with dictionary learning},
  author={Bricken, Trenton and Templeton, Adly and Batson, Joshua and Chen, Brian and Jermyn, Adam and Conerly, Tom and Turner, Nick and Anil, Cem and Denison, Carson and Askell, Amanda and others},
  journal={Transformer Circuits Thread},
  year={2023}
}

@article{gao2024scaling,
  title={Scaling and evaluating sparse autoencoders},
  author={Gao, Leo and la Tour, Tom Dupr{\'e} and Tillman, Henk and Goh, Gabriel and Troll, Rajan and Radford, Alec and Sutskever, Ilya and Leike, Jan and Wu, Jeff},
  journal={arXiv preprint arXiv:2406.04093},
  year={2024}
}

@article{templeton2024scaling,
  title={Scaling monosemanticity: Extracting interpretable features from {Claude} 3 {Sonnet}},
  author={Templeton, Adly and Conerly, Tom and Marcus, Jonathan and Lindsey, Jack and Bricken, Trenton and Chen, Brian and Pearce, Adam and Citro, Craig and Ameisen, Emmanuel and Jones, Andy and others},
  journal={Transformer Circuits Thread},
  year={2024}
}

@article{elhage2022superposition,
  title={Toy models of superposition},
  author={Elhage, Nelson and Hume, Tristan and Olsson, Catherine and Schiefer, Nicholas and Henighan, Tom and Kravec, Shaul and Hatfield-Dodds, Zac and Lasenby, Robert and Drain, Dawn and Chen, Carol and others},
  journal={Transformer Circuits Thread},
  year={2022}
}

@article{sharkey2022taking,
  title={Taking features out of superposition with sparse autoencoders},
  author={Sharkey, Lee and Braun, Dan and Millidge, Beren},
  journal={Alignment Forum},
  year={2022}
}

@article{theodoris2023transfer,
  title={Transfer learning enables predictions in network biology},
  author={Theodoris, Christina V and Xiao, Ling and Chopra, Anant and Chaffin, Mark D and Al Sayed, Zeina R and Hill, Matthew C and Manber, Helene and Engreitz, Jesse M and others},
  journal={Nature},
  volume={618},
  number={7965},
  pages={616--624},
  year={2023}
}

@article{cui2024scgpt,
  title={{scGPT}: toward building a foundation model for single-cell multi-omics using generative {AI}},
  author={Cui, Haotian and Wang, Chloe and Maan, Hassaan and Pang, Kuan and Luo, Fengning and Duan, Nan and Wang, Bo},
  journal={Nature Methods},
  volume={21},
  number={8},
  pages={1470--1480},
  year={2024}
}

@article{replogle2022mapping,
  title={Mapping information-rich genotype-phenotype landscapes with genome-scale {Perturb-seq}},
  author={Replogle, Joseph M and Saunders, Reuben A and Pogson, Alexandra N and Hussmann, Jeffrey A and Lenail, Alexander and Guna, Alina and Mascibroda, Lauren and Wagner, Eric J and Adelman, Karen and Lithwick-Yanai, Gila and others},
  journal={Cell},
  volume={185},
  number={14},
  pages={2559--2575},
  year={2022}
}

@article{han2018trrust,
  title={{TRRUST} v2: an expanded reference database of human and mouse transcriptional regulatory interactions},
  author={Han, Heonjong and Cho, Jae-Won and Lee, Sangyoung and Yun, Ayoung and Kim, Hyojin and Bae, Dasom and Yang, Sunmo and Kim, Chan Yeong and Lee, Minhyuk and Kim, Eunbeen and others},
  journal={Nucleic Acids Research},
  volume={46},
  number={D1},
  pages={D199--D205},
  year={2018}
}

@article{ashburner2000go,
  title={Gene {Ontology}: tool for the unification of biology},
  author={Ashburner, Michael and Ball, Catherine A and Blake, Judith A and Botstein, David and Butler, Heather and Cherry, J Michael and Davis, Allan P and Dolinski, Kara and Dwight, Selina S and Eppig, Janan T and others},
  journal={Nature Genetics},
  volume={25},
  number={1},
  pages={25--29},
  year={2000}
}

@article{kanehisa2000kegg,
  title={{KEGG}: {Kyoto} {Encyclopedia} of {Genes} and {Genomes}},
  author={Kanehisa, Minoru and Goto, Susumu},
  journal={Nucleic Acids Research},
  volume={28},
  number={1},
  pages={27--30},
  year={2000}
}

@article{jassal2020reactome,
  title={The {Reactome} pathway knowledgebase},
  author={Jassal, Bijay and Matthews, Lisa and Viteri, Guilherme and Gong, Chuqiao and Lorente, Pascual and Fabregat, Antonio and Sidiropoulos, Konstantinos and Cook, Justin and Gillespie, Marc and Haw, Robin and others},
  journal={Nucleic Acids Research},
  volume={48},
  number={D1},
  pages={D498--D503},
  year={2020}
}

@article{szklarczyk2023string,
  title={The {STRING} database in 2023: protein--protein association networks and functional enrichment analyses for any sequenced genome of interest},
  author={Szklarczyk, Damian and Kirsch, Rebecca and Koutrouli, Mikaela and Nastou, Katerina and Mehryary, Farrokh and Hachilif, Radja and Gable, Annika L and Fang, Tao and Doncheva, Nadezhda T and Pyysalo, Sampo and others},
  journal={Nucleic Acids Research},
  volume={51},
  number={D1},
  pages={D599--D610},
  year={2023}
}

@article{tabula2022tabula,
  title={The {Tabula Sapiens}: A multiple-organ, single-cell transcriptomic atlas of humans},
  author={{The Tabula Sapiens Consortium}},
  journal={Science},
  volume={376},
  number={6594},
  pages={eabl4896},
  year={2022}
}

@article{kendiukhov2025sae_atlas,
  title={Sparse autoencoders reveal organized biological knowledge but minimal regulatory logic in single-cell foundation models: A comparative atlas of {Geneformer} and {scGPT}},
  author={Kendiukhov, Ihor},
  journal={arXiv preprint},
  year={2025},
  note={Companion paper, Phases 1--3}
}

@article{kendiukhov2025systematic,
  title={Systematic evaluation of single-cell foundation model interpretability reveals attention captures co-expression rather than unique regulatory signal},
  author={Kendiukhov, Ihor},
  journal={arXiv preprint},
  year={2025},
  note={Companion paper}
}

@article{makhzani2013ksparse,
  title={k-Sparse autoencoders},
  author={Makhzani, Alireza and Frey, Brendan J},
  journal={arXiv preprint arXiv:1312.5663},
  year={2013}
}

@article{olshausen1997sparse,
  title={Sparse coding with an overcomplete basis set: A strategy employed by {V1}?},
  author={Olshausen, Bruno A and Field, David J},
  journal={Vision Research},
  volume={37},
  number={23},
  pages={3311--3325},
  year={1997}
}

@book{manning1999foundations,
  title={Foundations of Statistical Natural Language Processing},
  author={Manning, Christopher D and Sch{\"u}tze, Hinrich},
  year={1999},
  publisher={MIT Press}
}

@article{welford1962note,
  title={Note on a method for calculating corrected sums of squares and products},
  author={Welford, B. P.},
  journal={Technometrics},
  volume={4},
  number={3},
  pages={419--420},
  year={1962}
}

@book{cohen1988statistical,
  title={Statistical Power Analysis for the Behavioral Sciences},
  author={Cohen, Jacob},
  year={1988},
  edition={2nd},
  publisher={Lawrence Erlbaum Associates}
}

@article{bereska2024mechanistic,
  title={Mechanistic interpretability for {AI} safety---A review},
  author={Bereska, Leonard and Gavves, Efstratios},
  journal={arXiv preprint arXiv:2404.14082},
  year={2024}
}

@article{marks2024sparse,
  title={Sparse feature circuits: Discovering and editing interpretable causal graphs in language models},
  author={Marks, Samuel and Rager, Can and Michaud, Eric J. and Belinkov, Yonatan and Bau, David and Mueller, Aaron},
  journal={arXiv preprint arXiv:2403.19647},
  year={2024}
}

@inproceedings{wang2023interpretability,
  title={Interpretability in the wild: a circuit for indirect object identification in {GPT-2} small},
  author={Wang, Kevin and Variengien, Alexandre and Conmy, Arthur and Shlegeris, Buck and Steinhardt, Jacob},
  booktitle={International Conference on Learning Representations},
  year={2023}
}

@article{geiger2021causal,
  title={Causal abstractions of neural networks},
  author={Geiger, Atticus and Lu, Hanson and Icard, Thomas and Potts, Christopher},
  journal={Advances in Neural Information Processing Systems},
  volume={34},
  pages={9574--9586},
  year={2021}
}

@article{meng2022locating,
  title={Locating and editing factual associations in {GPT}},
  author={Meng, Kevin and Bau, David and Andonian, Alex and Belinkov, Yonatan},
  journal={Advances in Neural Information Processing Systems},
  volume={35},
  pages={17359--17372},
  year={2022}
}

@article{vig2020causal,
  title={Causal mediation analysis for interpreting neural {NLP}: The case of gender bias},
  author={Vig, Jesse and Gehrmann, Sebastian and Belinkov, Yonatan and Qian, Sharon and Nevo, Daniel and Sakenis, Simas and Huang, Jason and Singer, Yaron and Shieber, Stuart},
  journal={arXiv preprint arXiv:2004.12265},
  year={2020}
}

@article{ciccia2010ddr,
  title={The {DNA} damage response: making it safe to play with knives},
  author={Ciccia, Alberto and Elledge, Stephen J.},
  journal={Molecular Cell},
  volume={40},
  number={2},
  pages={179--204},
  year={2010}
}

@article{jackson2009ddr,
  title={The {DNA}-damage response in human biology and disease},
  author={Jackson, Stephen P. and Bartek, Jiri},
  journal={Nature},
  volume={461},
  number={7267},
  pages={1071--1078},
  year={2009}
}

@article{malumbres2009cell,
  title={Cell cycle, {CDKs} and cancer: a changing paradigm},
  author={Malumbres, Marcos and Barbacid, Mariano},
  journal={Nature Reviews Cancer},
  volume={9},
  number={3},
  pages={153--166},
  year={2009}
}

@article{musacchio2007spindle,
  title={The spindle-assembly checkpoint in space and time},
  author={Musacchio, Andrea and Salmon, Edward D.},
  journal={Nature Reviews Molecular Cell Biology},
  volume={8},
  number={5},
  pages={379--393},
  year={2007}
}

@article{labbadia2015proteostasis,
  title={The biology of proteostasis in aging and disease},
  author={Labbadia, Johnathan and Morimoto, Richard I.},
  journal={Annual Review of Biochemistry},
  volume={84},
  pages={435--464},
  year={2015}
}

@article{hetz2012unfolded,
  title={The unfolded protein response: controlling cell fate decisions under {ER} stress and beyond},
  author={Hetz, Claudio},
  journal={Nature Reviews Molecular Cell Biology},
  volume={13},
  number={2},
  pages={89--102},
  year={2012}
}

@article{walter2011unfolded,
  title={The unfolded protein response: from stress pathway to homeostatic regulation},
  author={Walter, Peter and Ron, David},
  journal={Science},
  volume={334},
  number={6059},
  pages={1081--1086},
  year={2011}
}

@article{nusse2017wnt,
  title={{Wnt}/\ensuremath{\beta}-catenin signaling, disease, and emerging therapeutic modalities},
  author={Nusse, Roel and Clevers, Hans},
  journal={Cell},
  volume={169},
  number={6},
  pages={985--999},
  year={2017}
}

@article{dhillon2007map,
  title={{MAP} kinase signalling pathways in cancer},
  author={Dhillon, Amardeep S. and Hagan, Suzanne and Rath, Olga and Kolch, Walter},
  journal={Oncogene},
  volume={26},
  number={22},
  pages={3279--3290},
  year={2007}
}

@article{boulon2010nucleolus,
  title={The nucleolus under stress},
  author={Boulon, S{\'e}bastien and Westman, Belinda J. and Hutten, Saskia and Boisvert, Fran{\c{c}}ois-Michel and Lamond, Angus I.},
  journal={Molecular Cell},
  volume={40},
  number={2},
  pages={216--227},
  year={2010}
}

@article{simons2000cholesterol,
  title={How cells handle cholesterol},
  author={Simons, Kai and Ikonen, Elina},
  journal={Science},
  volume={290},
  number={5497},
  pages={1721--1726},
  year={2000}
}

@article{chandel2021mitochondria,
  title={Mitochondria},
  author={Chandel, Navdeep S.},
  journal={Cold Spring Harbor Perspectives in Biology},
  volume={13},
  number={3},
  pages={a040584},
  year={2021}
}

@article{yang2022scbert,
  title={{scBERT} as a large-scale pretrained deep language model for cell type annotation of single-cell {RNA-seq} data},
  author={Yang, Fan and Wang, Wenchuan and Wang, Fang and Fang, Yuan and Tang, Duyu and Huang, Junzhou and Lu, Hui and Yao, Jianhua},
  journal={Nature Machine Intelligence},
  volume={4},
  number={10},
  pages={852--866},
  year={2022}
}

@article{hao2024large,
  title={Large scale foundation model on single-cell transcriptomics},
  author={Hao, Minsheng and Gong, Jing and Zeng, Xin and Liu, Chiming and Guo, Yucheng and Cheng, Xingyi and Wang, Taifeng and Ma, Jianzhu and Zhang, Xuegong and Song, Le},
  journal={Nature Methods},
  volume={21},
  number={8},
  pages={1481--1491},
  year={2024}
}

@article{ross2004proteostasis,
  title={Protein aggregation and neurodegenerative disease},
  author={Ross, Christopher A. and Poirier, Michelle A.},
  journal={Nature Medicine},
  volume={10},
  number={Suppl},
  pages={S10--S17},
  year={2004}
}
